\documentclass[runningheads]{llncs}

\usepackage[mobile]{eccv}

\usepackage{eccvabbrv}

\usepackage{graphicx}
\usepackage{booktabs}
\usepackage{marvosym}
\usepackage[linesnumbered,ruled]{algorithm2e}
\usepackage[accsupp]{axessibility}

\usepackage[pagebackref=false,breaklinks=true,colorlinks,citecolor=eccvblue,bookmarks=true,bookmarksnumbered=true]{hyperref}
\hypersetup{
  pdftitle={GlobalPointer: Large-Scale Plane Adjustment with Bi-Convex Relaxation},
  pdfsubject={Computer Vision, Robotics},
  pdfauthor={Bangyan Liao, Zhenjun Zhao, Lu Chen, Haoang Li, Daniel Cremers, Peidong Liu},
  pdfkeywords={Plane Adjustment, Semidefinite Programming(SDP), Convex Relaxation}
}
\usepackage[absolute]{textpos}

\usepackage{orcidlink}

\newcommand{\PAR}[1]{\vspace{0.1cm}\noindent{\bf #1} }

\begin{document}

\title{GlobalPointer: Large-Scale Plane Adjustment with Bi-Convex Relaxation}

\definecolor{somegray}{gray}{0.5}
\newcommand{\darkgrayed}[1]{\textcolor{somegray}{#1}}
\begin{textblock}{11}(2.5, -0.1)  %
\begin{center}
\darkgrayed{This paper has been accepted for publication at the European Conference on Computer Vision (ECCV), 2024. \copyright Springer}
\end{center}
\end{textblock}

\titlerunning{GlobalPointer}

\author{
Bangyan Liao\inst{1,2\star}\orcidlink{0009-0007-7739-4879} \and
Zhenjun Zhao\inst{3\star}\orcidlink{0009-0000-6551-4537} \and
Lu Chen\inst{4}\orcidlink{0009-0003-5779-4673} \and
Haoang Li\inst{5}\orcidlink{0000-0002-1576-9408}\and\\
Daniel Cremers\inst{6}\orcidlink{0000-0002-3079-7984}\and
Peidong Liu\inst{2}\textsuperscript{\Letter}\orcidlink{0000-0002-9767-6220}
}

\authorrunning{B. Liao, Z. Zhao \etal}

\institute{\textsuperscript{1}Zhejiang University \quad \textsuperscript{2}Westlake University \\ \textsuperscript{3}The Chinese University of Hong Kong \quad \textsuperscript{4}Dreame Technology (Suzhou) \\ \textsuperscript{5}Hong Kong University of Science and Technology (Guangzhou) \\ \textsuperscript{6}Technical University of Munich}

\maketitle

\let\thefootnote\relax\footnotetext{$^\star$ Equal contribution.} 
\footnotetext{$^\text{\Letter}$ Corresponding author: Peidong Liu (liupeidong@westlake.edu.cn).}

\begin{abstract}
Plane adjustment (PA) is crucial for many 3D applications, involving simultaneous pose estimation and plane recovery. Despite recent advancements, it remains a challenging problem in the realm of multi-view point cloud registration. Current state-of-the-art methods can achieve globally optimal convergence only with good initialization. Furthermore, their high time complexity renders them impractical for large-scale problems. To address these challenges, we first exploit a novel optimization strategy termed \textit{Bi-Convex Relaxation}, which decouples the original problem into two simpler sub-problems, reformulates each sub-problem using a convex relaxation technique, and alternately solves each one until the original problem converges. Building on this strategy, we propose two algorithmic variants for solving the plane adjustment problem, namely \textit{GlobalPointer} and \textit{GlobalPointer++}, based on point-to-plane and plane-to-plane errors, respectively. Extensive experiments on both synthetic and real datasets demonstrate that our method can perform large-scale plane adjustment with linear time complexity, larger convergence region, and robustness to poor initialization, while achieving similar accuracy as prior methods. The code is available at \href{https://github.com/wu-cvgl/GlobalPointer}{github.com/wu-cvgl/GlobalPointer}.
	
\end{abstract}
\section{Introduction}

With the widespread adoption of LiDAR technology in applications such as 3D reconstruction and LiDAR SLAM \cite{zhang2014loam, shan2018lego}, tasks involving localization and scene modeling have gained significant attention. As a fundamental building block, there is an increasing demand for a more efficient, robust, and accurate multi-frame point cloud registration algorithm for downstream tasks.

Although the two-frame point cloud registration problem \cite{yang2020teaser, segal2009generalized, briales2017convex} has been extensively studied in the computer vision community for decades, transitioning to multi-frame scenarios introduces new challenges. Specifically, the relative poses obtained from pairwise point cloud registration often result in the well-known pose drift problem \cite{BALM2}. Many efforts, such as pose graph optimization \cite{rosen2019se} and rotation averaging \cite{Zhang_2023_CVPR, chen2021hybrid}, have been made to address this challenge by introducing additional pose observations to average out pose errors. However, these methods often yield sub-optimal and biased results due to the complex pose noise models.

\begin{figure}[tb]
	\centering
	\begin{subfigure}[b]{0.33\textwidth}
		\centering
		\includegraphics[width=\textwidth]{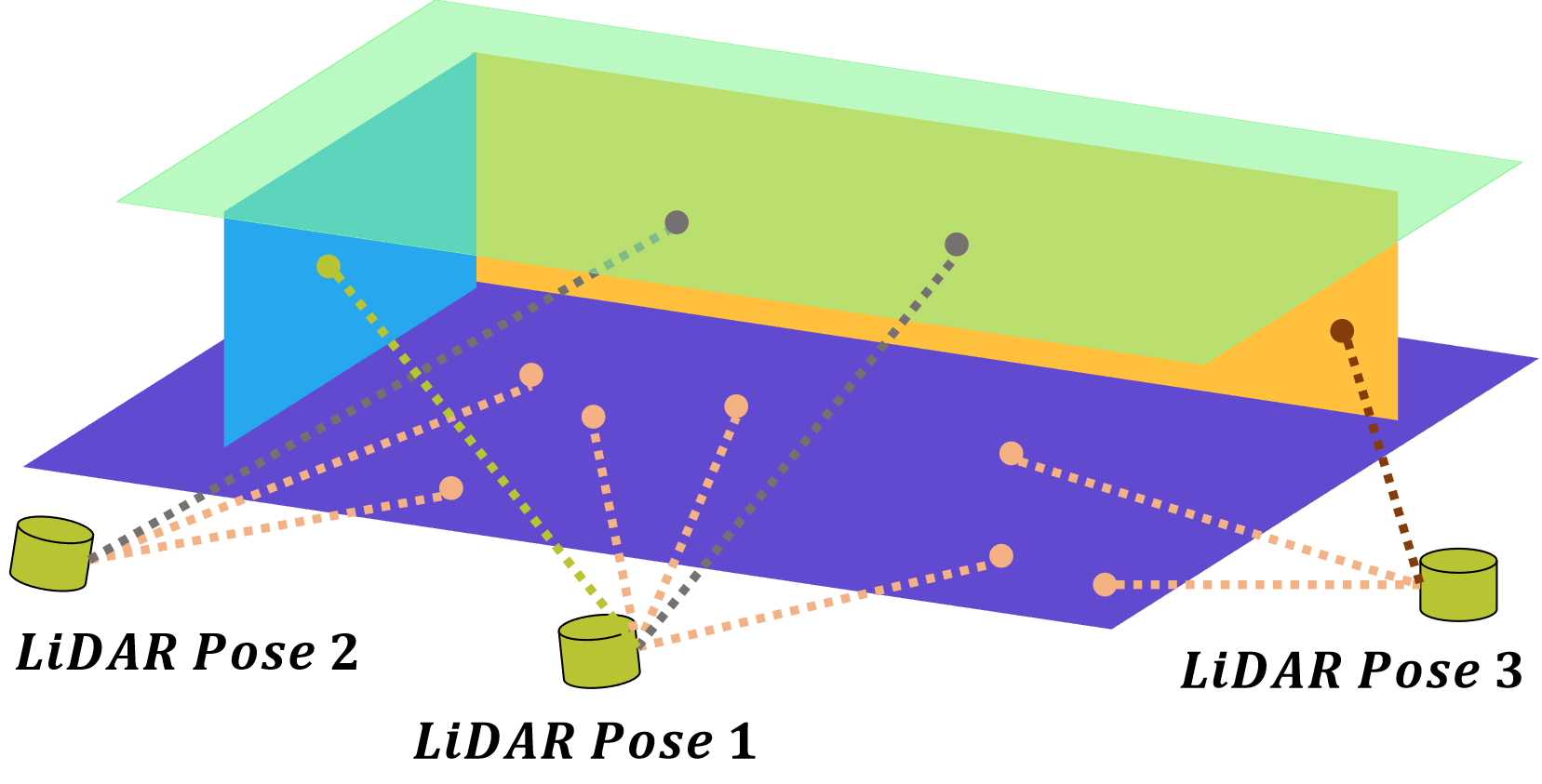}
		\caption{}
		\label{fig:front1}
	\end{subfigure}
	\hfill
	\begin{subfigure}[b]{0.60\textwidth}
		\centering
		\includegraphics[width=\textwidth]{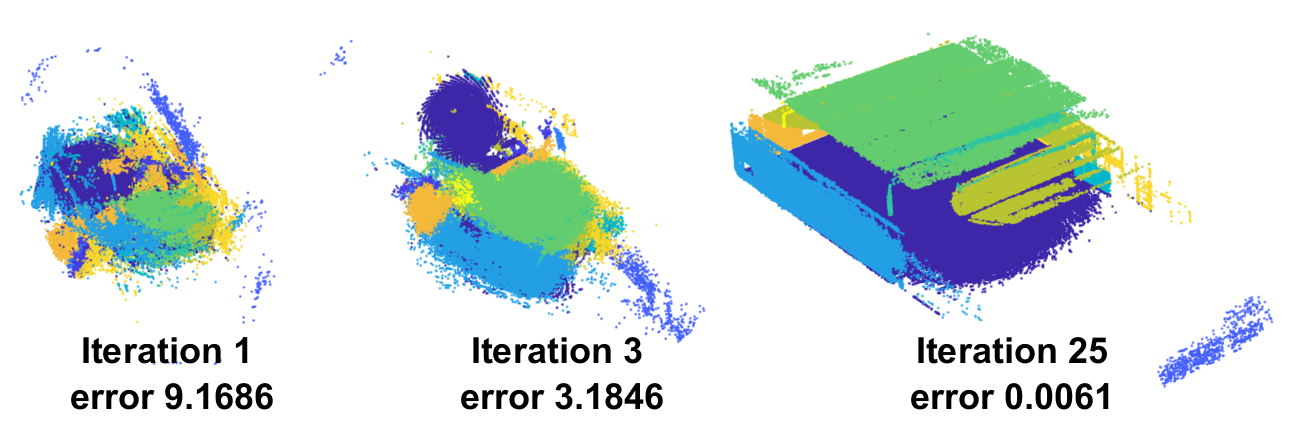}
		\caption{}
		\label{fig:front3}
	\end{subfigure}
    \caption{\textbf{Visualization of plane adjustment problem and convergence process.} (a) The plane adjustment problem involves simultaneous pose estimation and plane reconstruction based on the assigned plane labels of each LiDAR point. As illustrated, the LiDAR poses and the 3D planes are tightly coupled with each other. (b) This example demonstrates the convergence process of our method starting from random initialization. Despite the poses and planes being randomly initialized, our approach achieves a global minimum within a few optimization iterations. For visualization purposes, points associated with the same plane are colored identically.}
	\label{fig_teaser}
\end{figure}

Inspired by the success of bundle adjustment \cite{triggs2000bundle}, researchers have explored its replication in multi-frame point cloud registration, leading to the recent development of plane adjustment techniques \cite{hsiao2017keyframe, kaess2015simultaneous}. Similar to bundle adjustment, plane adjustment simultaneously optimizes the camera poses and plane parameters (as the counterpart of 3D points in bundle adjustment) by minimizing point-to-plane \cite{zhou2020efficient} or plane-to-plane \cite{kaess2015simultaneous} errors, as shown in \cref{fig:front1}.

Prior state-of-the-art methods explore both nonlinear least square methods \cite{pi-SLAM, zhou2020efficient} and spectral-based methods \cite{BALM,BALM2,EF,EF2,ESO} to improve the performance of plane adjustment.
While existing works on plane adjustment have shown promising results,  these methods are limited to small-scale problems, and their accuracy depends on the quality of initialization. Achieving both large-scale and globally optimal plane adjustment remains challenging.

To address these challenges, we exploit a novel optimization strategy termed \textit{Bi-Convex Relaxation} for the large-scale plane adjustment problem. This strategy decouples the original complex formulation into two sub-problems. Each sub-problem is reformulated using convex relaxation techniques \cite{anstreicher2012convex} and solved alternately until the overall problem converges. The advantages of our method are two-fold: 1) the convex sub-problem enlarges the convergence region, enhancing robustness to poor initialization; and 2) decoupling the high-dimensional problem into multiple low-dimensional sub-problems avoids solving intractable large Semidefinite Programming (SDP) problem, enabling efficient optimization for large-scale scenarios. Building upon this framework, we present two algorithmic variants, namely \textit{GlobalPointer} and \textit{GlobalPointer++}, based on point-to-plane and plane-to-plane errors, respectively. The former algorithm exhibits a larger convergence region and better stability, while the latter demonstrates superior efficiency. Although there is no theoretical globally optimal guarantee, exhaustive empirical evaluations in both synthetic and real experiments demonstrate that our method can provide an empirical globally optimal solution, as shown in \cref{fig:front3}. 

In summary, our {\bf{contributions}} are as follows:
\begin{itemize}
	\item We exploit a novel optimization strategy termed \textit{Bi-Convex Relaxation}, which combines the advantages of both alternating minimization \cite{wang2008new} and convex relaxation techniques \cite{anstreicher2012convex};
	\item Building on this novel optimization strategy, we develop two algorithmic variants for plane adjustment, namely \textit{GlobalPointer} and \textit{GlobalPointer++}, which depend on point-to-plane and plane-to-plane errors, respectively;
	\item Extensive synthetic and real experimental evaluations demonstrate that our method can perform large-scale plane adjustment with linear time complexity and robustness to poor initialization, while achieving similar accuracy as prior methods.
\end{itemize}
\section{Related Work}

Kaess \etal \cite{kaess2015simultaneous} exploiting plane-to-plane error to formulate the plane adjustment problem as a nonlinear least squares problem, which they solved efficiently using the Gauss-Newton optimization method. Hsiao \etal~\cite{hsiao2017keyframe} extend this approach to a keyframe-based SLAM system. Zhou \etal~\cite{zhou2020efficient} later demonstrate that using point-to-plane error in the energy function formulation improves stability and efficiency over plane-to-plane error formulations. They utilize the matrix factorization trick to solve the resulting nonlinear least squares problem, avoiding the accumulation of a large number of point clouds. This method has been widely adopted in plane-based SLAM systems \cite{pi-SLAM}.

While these formulations are very efficient for small-scale problems (\ie, with a small number of camera poses and planes), they struggle in large-scale problems due to the inherent time complexity. To address this issue, Ferrer \etal~\cite{EF} propose to use the minimum eigenvalue of the covariance matrix to obtain a surrogate energy function. This approach avoids the explicit plane updates and requires only eigenvalue decomposition at each iteration. Analytical gradients are derived, and a first-order solver is used for solving this problem. To further improve the convergence speed, they derive the analytical Hessian matrix, enabling more efficient optimization with a second-order solver \cite{EF2}. Similarly, Liu \etal derive a similar analytical Hessian matrix and further improve overall efficiency in \cite{BALM,BALM2}. While methods based on eigenvalue decomposition can avoid explicit plane parameter updates, constructing the Hessian matrix itself is time-consuming, and performing eigenvalue decomposition at each iteration further increases the computational burden. Recently, Zhou \cite{ESO} proposes a novel approach to exploit implicit constraints of eigenvalues to derive analytical Hessian matrices and gradient vectors. However, direct implicit function differentiation would potentially pose a numerical stability issue.

More recently, the convex relaxation technique has been increasingly employed to solve challenging non-convex optimization problems in computer vision tasks \cite{yang2020teaser,RCD,briales2017convex,dellaert2020shonan}. Nevertheless, the efficiency of SDP, as a core computational tool, is highly related to the size of the state matrix \cite{wolkowicz2012handbook}, making it intractable for large-scale plane adjustment.

To address these challenges, we propose to hybridize two techniques, \ie, alternating minimization \cite{wang2008new} and convex relaxation \cite{anstreicher2012convex}, to decouple the original complex plane adjustment problem into two simpler sub-problems. This technique, termed \textit{Bi-Convex Relaxation}, avoids solving high-dimensional SDP problem, significantly reduces time complexity, and enlarges the convergence region. Although there is no theoretical guarantee for the entire problem, our empirical results indicate that our formulation can converge to a global minimum.
\section{Notations and Background}

\subsection{Notation}
We use MATLAB notation to denote sub-matrix operations. Specifically, $[\mathbf{a};\mathbf{b}]$ denotes the vertical concatenation of vectors of $\mathbf{a}$ and $\mathbf{b}$, while $[\mathbf{a}^{\top},\mathbf{b}^{\top}]$ denotes the horizontal concatenation of the transpose vectors of $\mathbf{a}$ and $\mathbf{b}$. The operator $\otimes$ denotes the Kronecker product, and $\times$ denotes the cross product of vectors. The operator $vec(.)$ denotes vertical vectorization. For the efficiency of the algorithm, we use a unit quaternion parameterized rotation matrix in some tasks.

\subsection{Point-to-Plane Error}
\label{sec:point2plane_error}

Given the rotation matrix $\mathbf{R} \in \mathbf{SO(3)}$ and translation vector $\mathbf{t} \in \mathbb{R}^3$ , we can transform each 3D point from local coordinates $\mathbf{P}^l\in \mathbb{R}^3$ to world coordinates $\mathbf{P}^g \in \mathbb{R}^3$ as $\mathbf{P}^g = \mathbf{R}\mathbf{P}^l + \mathbf{t}$.

\begin{figure}[tb]
	\centering
	\begin{subfigure}[b]{0.44\textwidth}
		\centering
		\includegraphics[width=\textwidth]{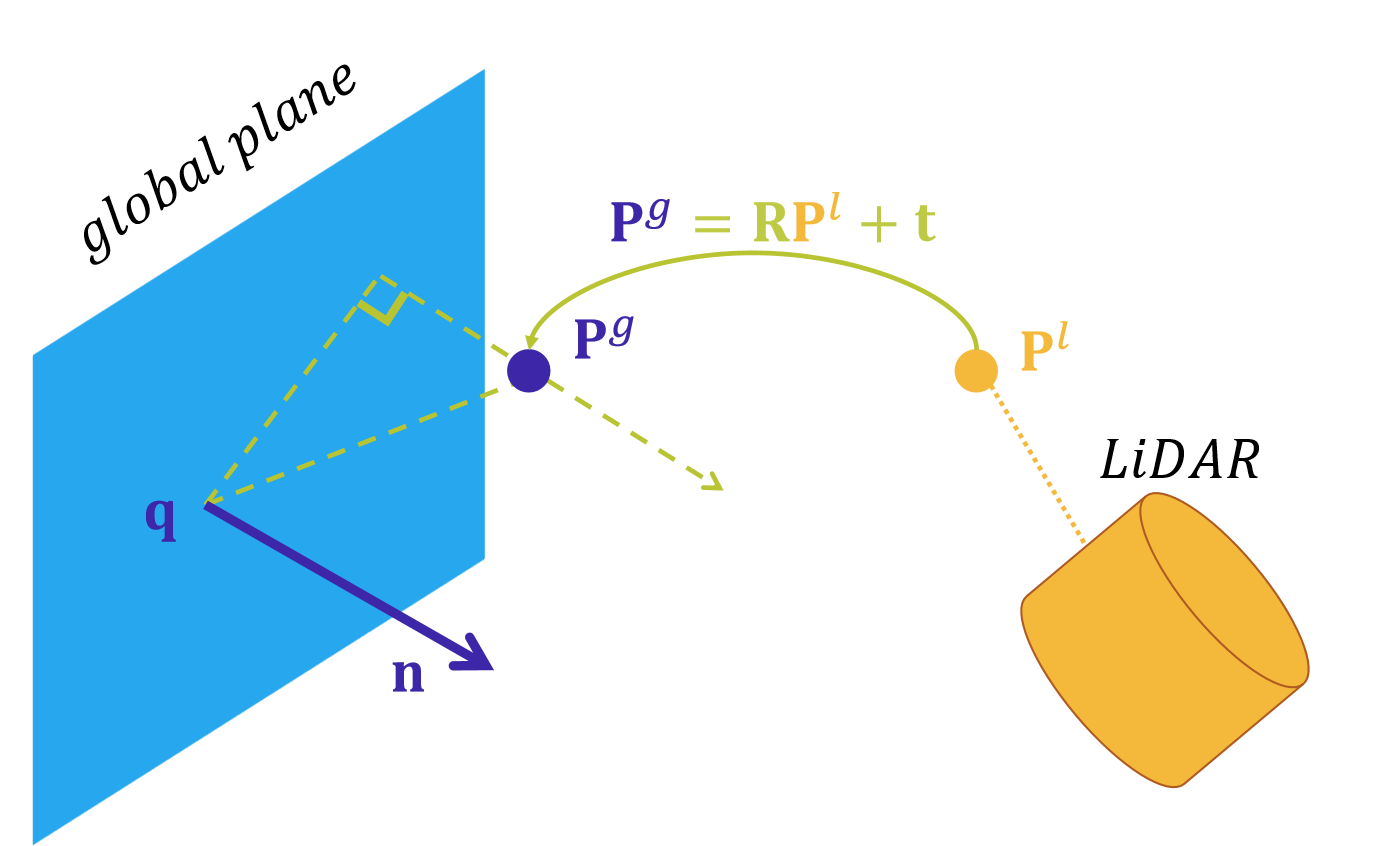}
		\caption{Point-to-Plane Error Definition.}
		\label{fig:error_def1}
	\end{subfigure}
	\hfill
	\begin{subfigure}[b]{0.54\textwidth}
		\centering
		\includegraphics[width=\textwidth]{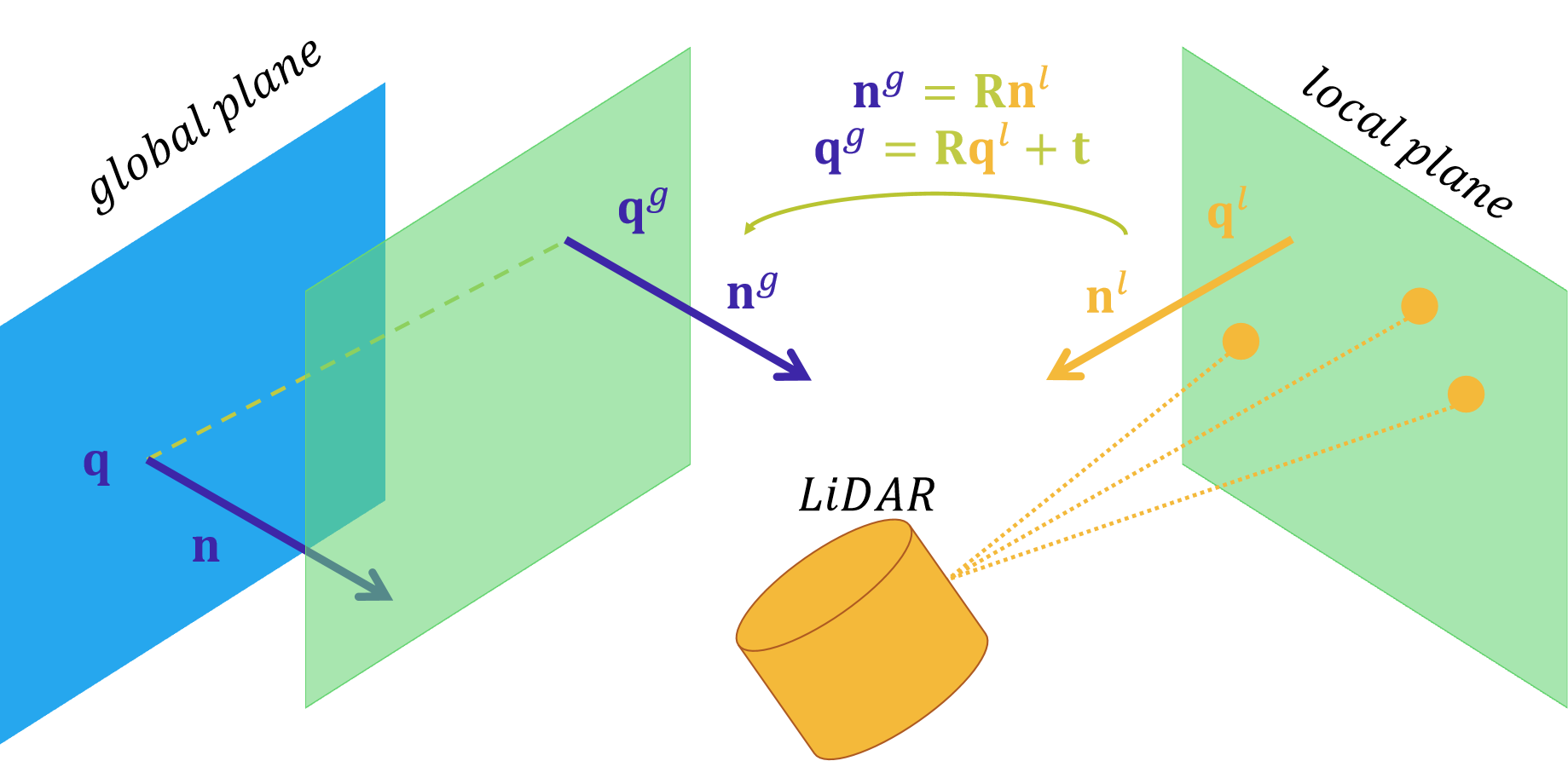}
		\caption{Plane-to-Plane Error Definition}
		\label{fig:error_def2}
	\end{subfigure}
	\caption{\textbf{Visualization of different point cloud error definitions.}}
	\label{fig:error_def}
\end{figure}

For each $\mathbf{P}^g$, the corresponding plane parameters $\mathbf{n}\in \mathbb{R}^3, \mathbf{q}\in \mathbb{R}^3$ are defined as the normal vector and an arbitrary point on the plane, respectively. To simplify, we introduce an auxiliary scalar $d=-\mathbf{n}^{\top}\mathbf{q}$. We then establish a point-to-plane distance as a metric for measuring the registration error, illustrated in \cref{fig:error_def1}, as ${\left \| \mathbf{n}^{\top}  (\mathbf{P}^g - \mathbf{q})\right \|}_2^2={\left \| \mathbf{n}^{\top}  \mathbf{P}^g + d\right \|}_2^2$.

\subsection{Plane-to-Plane Error}
\label{sec:plane2plane_error}
Similar to the point-to-plane error, we can define the plane-to-plane error. Since LiDAR point clouds are typically dense, we can always find many LiDAR points in a single frame that correspond to the same 3D plane. We can use plane segmentation and fitting algorithms to obtain the parameters of these planes $\mathbf{n}_k^l$ and $\mathbf{q}_k^l$, in the local LiDAR coordinate system. Given the rotation matrix $\mathbf{R}$, translation vector $\mathbf{t}$, normal vector $\mathbf{n}^l$, and an arbitrary point $\mathbf{q}^l$, we can readily transform them to the world coordinate as $\mathbf{n}^g = \mathbf{R}\mathbf{n}^l$ and $\mathbf{q}^g = \mathbf{R}\mathbf{q}^l + \mathbf{t}$, respectively.
Given the corresponding global plane parameters $\mathbf{n}$ and $\mathbf{q}$, we can establish a plane-to-plane distance as a metric for measuring the registration error, illustrated in \cref{fig:error_def2}, as ${\left \| \mathbf{n}^g - \mathbf{n} \right \|}_2^2 + {\left \|\mathbf{n}^{g\top}\mathbf{q}^g  -  \mathbf{n}^{\top}\mathbf{q} \right \|}_2^2={\left \| \mathbf{n}^g - \mathbf{n} \right \|}_2^2 + {\left \|d  -  d^g \right \|}_2^2$. 

\subsection{Convex Relaxation}
Quadratically Constrained Quadratic Program (QCQP), as a general but NP-hard optimization problem, has various applications across computer vision and machine learning \cite{zhao2020efficient}. It can be defined as:
\begin{equation}
    \begin{aligned}
        \min_{\mathbf{x} \in \mathbb{R}^n} \quad & \mathbf{x}^{\top} \mathbf{C} \mathbf{x}\\
         \text{s.t. } \quad & \mathbf{x}^{\top} \mathbf{A}_i \mathbf{x} = b_i, \quad i = 1, \ldots, m,
    \end{aligned}
\end{equation}
where $\mathbf{C}, \mathbf{A}_1,...,\mathbf{A}_m \in \mathcal{S}^n$ and $\mathcal{S}^n$ denotes the set of all real symmetric $n \times n$ matrices.

The convex relaxation technique serves as a general tool to reformulate the original QCQP problem into a new convex problem, allowing it to be solved to a global minimum. Specifically, we first rewrite $\mathbf{x}^{\top} \mathbf{C} \mathbf{x}$ as $trace(\mathbf{C} \mathbf{x}\mathbf{x}^{\top} )$ and subsequently replace $\mathbf{x}\mathbf{x}^{\top}$ with a new symmetric positive semidefinite (PSD) matrix $\mathbf{X}$. This leads to a new Semidefinite Programming (SDP) problem:
\begin{equation}
    \begin{aligned}
        \min_{\mathbf{X} \in \mathcal{S}^n} \quad & trace(\mathbf{C}\mathbf{X})\\
         \text{s.t. } \quad & trace(\mathbf{A}_i\mathbf{X}) = b_i, \quad i = 1, \ldots, m, \\
         & \mathbf{X} \succeq \mathbf{0}.
            \end{aligned}
\end{equation}

The primary benefit of employing convex relaxation instead of directly solving the original problem lies in the convex nature of the SDP formulation. Although relaxation is applied, researchers have found that strong duality properties still hold for many problems \cite{rosen2019se,harenstam2023semidefinite,zhao2020efficient,briales2017convex,eriksson2018rotation}. This implies that solving the relaxed problem is often equivalent to solving the original problem \cite{anstreicher2012convex}. With any off-the-shelf SDP solver \cite{MOSEK}, we can always find the global minimum within a polynomial time.

\section{Methodology}
In this section, we first define the formal plane adjustment algorithm in \cref{subsec:Plane Adjustment}. Then, a new optimization strategy called \textit{Bi-Convex Relaxation} will be established in \cref{subsec:Bi-Convex Relaxation}. Building on this strategy, we propose the \textit{GlobalPointer} algorithm, based on the point-to-plane error, in \cref{subsec:GlobalPointer}. In \cref{subsec:GlobalPointer++}, we accelerate the original algorithm with two closed-form solvers in \textit{GlobalPointer++} based on the plane-to-plane error. We further incorporate some empirical insights and intuitive remarks on our solvers in \cref{subsec:Performance Analysis}. 

\subsection{Plane Adjustment (PA)}
\label{subsec:Plane Adjustment}
Let us consider $m$ LiDAR frames and $n$ reconstructed planes. For the $i^{th} (i \in \{1,2,\dots,m\})$ LiDAR frame, we define its absolute pose in the global world coordinates by a rotation matrix $\mathbf{R}_i \in \mathbf{SO(3)}$ and a translation vector $\mathbf{t}_i \in \mathbb{R}^3$. For the $j^{th} (j \in \{ 1,2,\dots,n\})$ reconstructed plane, we define $\mathbf{n}_j$ as its normal vector and $\mathbf{q}_j$ as its origin point.

\PAR{PA with Point-to-Plane Error.}
When utilizing the point-to-plane error in \cref{sec:point2plane_error}, the plane adjustment problem with point-to-plane error can be defined as:
\begin{equation}
	\begin{aligned}
		\{\mathbf{R}^{*} , \mathbf{t}^{*}, \mathbf{n}^{*}, \mathbf{q}^{*}\} = arg \min_{\mathbf{R}, \mathbf{t}, \mathbf{n}, \mathbf{q}} \quad & \sum_{i = 1}^{m} \sum_{j = 1}^{n} {\begin{bmatrix}
				\mathbf{n}_j \\
				d_j
		\end{bmatrix}}^{\top}\mathbf{T}_i \mathcal{B}(i, j)\mathbf{T}_i^{\top}\begin{bmatrix}
			\mathbf{n}_j \\
			d_j 
		\end{bmatrix}\\
		\text{s.t. } \quad 
        & \mathbf{R}_i \in  \mathbf{SO(3)}, \quad \left \| \mathbf{n}_j \right \| = 1, \\
        & \mathcal{B}(i, j) = \sum_{k \in obs(i, j)}(\tilde{\mathbf{P}}_k^l \tilde{\mathbf{P}}_k^{l\top}), \\
		& i = 1, \ldots, m, \quad j = 1, \ldots, n,
	\end{aligned}
	\label{eq:point-to-plane PA}
\end{equation}
where $obs(i, j)$ denotes the observation index set related to the $j^{th}$ reconstructed plane and the $i^{th}$ LiDAR frame, $\mathbf{T}_i = [\mathbf{R}_i, \mathbf{t}_i; \mathbf{0} ,1]$ represents the transformation matrix for LiDAR frame $i$, and $\tilde{\mathbf{P}}_k^l = [\mathbf{P}_k^l;1]$ denotes homogeneous coordinates of point $\mathbf{P}_k^l$ .

{\it Remark.} The matrix $\mathcal{B}(i, j)$ accumulates all relevant local points in advance, avoiding time-consuming point cloud accumulation operations. Nonetheless, solving this problem is still hard due to the non-convex nature of both the objective function and constraints, making it highly reliant on a good initialization.

\subsubsection{PA with Plane-to-Plane Error.}\label{subsubsection:PA with Plane-to-Plane} 
When utilizing the plane-to-plane error in \cref{sec:plane2plane_error}, the plane adjustment problem with plane-to-plane error can be defined as:
\begin{equation}
	\begin{aligned}
		\{\mathbf{R}^{*} , \mathbf{t}^{*}, \mathbf{n}^{*}, \mathbf{q}^{*}\} = arg \min_{\mathbf{R}, \mathbf{t}, \mathbf{n}, \mathbf{q}} \quad & \sum_{i = 1}^{m} \sum_{j = 1}^{n} {\left \| \mathbf{n}_j - \mathbf{R}_i\mathbf{n}_{ij}^{l*} \right \|}_2^2 + {\left \| \mathbf{n}_{ij}^{l*\top}\mathbf{R}_i^{\top}\mathbf{t}_i  + d_j - {d}_{ij}^{l*}  \right \|}_2^2  \\
		\text{s.t. } \quad 
        & \mathbf{R}_i \in  \mathbf{SO(3)}, \quad \left \| \mathbf{n}_j \right \| = 1, \\
		& i = 1, \ldots, m, \quad j = 1, \ldots, n,
	\end{aligned}
	\label{eq:plane-to-plane PA}
\end{equation}
where the optimal local plane parameters $[\mathbf{n}_{ij}^{l*}; {d}_{ij}^{l*}]$ is obtained by minimizing the point-to-plane error for LiDAR points observed in the $i^{th}$ LiDAR frame associated with the $j^{th}$ reconstructed plane as:
\begin{equation}
    (\mathbf{n}_{ij}^{l*}, {d}_{ij}^{l*}) = arg \min_{\mathbf{n}_{ij}^l, {d}_{ij}^l}  \begin{bmatrix}
		\mathbf{n}_{ij}^l\\{d}_{ij}^l
		\end{bmatrix}^{\top}\mathcal{B}(i, j)\begin{bmatrix}
		    \mathbf{n}_{ij}^l\\{d}_{ij}^l
		\end{bmatrix}.
\end{equation}

{\it Remark.} As detailed in the subsequent section, the formulation akin to a nonlinear least squares problem simplifies \cref{eq:plane-to-plane PA}. However, this also comes at a price. The direction of the normal vector for each local plane introduces ambiguity. Failure to unify these directions can lead to divergence in the entire optimization process.

\begin{figure}[tb]
	\centering
	\includegraphics[height=3cm]{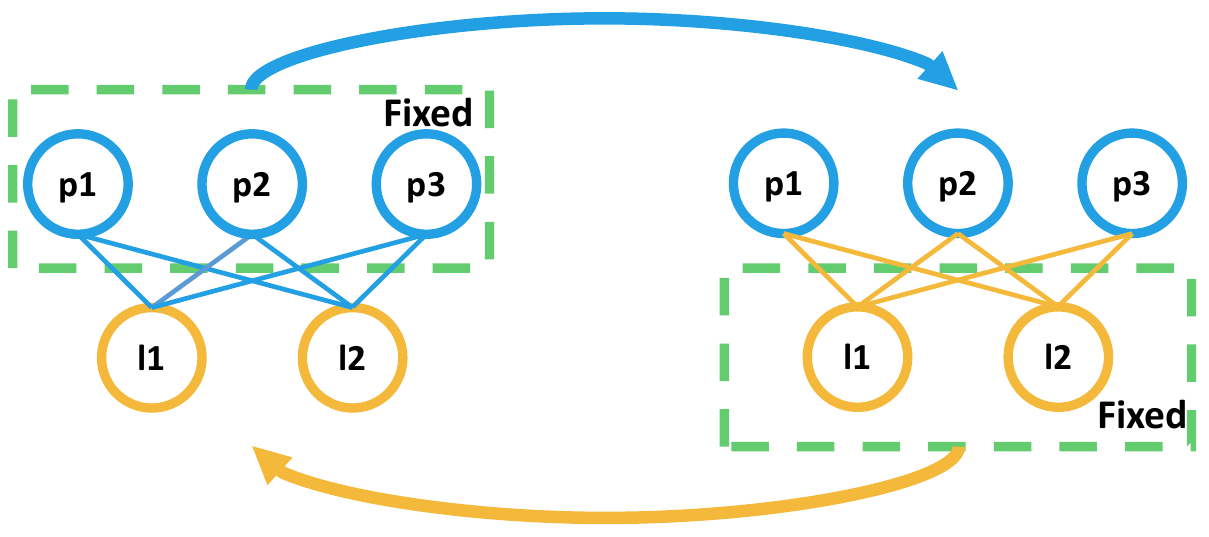}
	\caption{\textbf{Illustration of the proposed \textit{Bi-Convex Relaxation}}. Three blue circles represent plane1, plane2, and plane3. Two yellow circles denote the LiDAR pose1 and LiDAR pose2. The connection edges denote the observation relationships. Fixing the planes leads to pose-only optimization (left), while fixing the LiDAR poses in plane-only optimization (right). Each sub-problem is then reformulated as SDP, which can be solved with globally optimal guarantees. Using these techniques, the plane adjustment algorithm alternately solves each sub-problem until the convergence of the original problem.}
	\label{fig:decouple_optimization}
\end{figure}

\subsection{Bi-Convex Relaxation}
\label{subsec:Bi-Convex Relaxation}
As shown above, whether based on point-to-plane or plane-to-plane error, direct plane adjustment presents a non-convex optimization problem. Fortunately, when either the planes or the poses are fixed, the remaining optimization problem becomes a QCQP problem, which can be reformulated with convex relaxation and solved to the global minimum. Thus, we adopt an alternating minimization approach to achieve convergence in the original plane adjustment. We term this strategy, which combines alternating minimization and convex relaxation, as \textit{Bi-Convex Relaxation}, shown in \cref{fig:decouple_optimization}. Although the non-convex nature of the original problem prevents \textit{Bi-Convex Relaxation} from guaranteeing global convergence, extensive experiments demonstrate that this algorithm significantly enlarges the convergence region and even provides empirical global optimality guarantees under certain conditions. Additionally, fixing a subset of parameters decouples the inter-dependence among the remaining parameters for optimization (\eg fixing $p1$, $p2$, $p3$ makes $l1$, $l2$ independent in \cref{fig:decouple_optimization}). This allows for parallel optimization of each plane or pose, avoiding high-dimensional SDP optimization and significantly reducing time complexity.
\begin{algorithm}[t]
    \SetKwInOut{Input}{Input}
    \SetKwInOut{Output}{Output}
    \Input{Random initialized LiDAR poses $\{\mathbf{R}_i,\mathbf{t}_i\mid i=1,\dots,m\}$, random initialized plane parameters $\{\mathbf{n}_j,\mathbf{q}_j\mid j=1,\dots,n\}$, and labeled LiDAR points $\{\mathbf{P}_k\mid k\in obs(i,j)\}$.}
    \Output{Optimized LiDAR poses and plane parameters $\{\mathbf{R}^*,\mathbf{t}^*,\mathbf{n}^*,\mathbf{q}^*\}$.}
    
    \While{(not reach max iteration) and (not satisfy stopping criteria)}
    {
        Find $\{\mathbf{X}^*\}$ by solving the \ref{eq:pose_only_sdp}\;
        Rounding $\{\mathbf{X}^*\}$ and recover LiDAR poses\;
        Find $\{\mathbf{Y}^*\}$ by solving the \ref{eq:plane_only_sdp}\;
        Rounding $\{\mathbf{Y}^*\}$ and recover plane parameters\;
    }
    \caption{\textit{GlobalPointer}}
    \label{alg:global_pointer}
\end{algorithm}

\subsection{GlobalPointer}
\label{subsec:GlobalPointer}
In this subsection, we will follow the \textit{Bi-Convex Relaxation} technique to derive the SDP formulation for each sub-problem based on point-to-plane error. The algorithm is summarized in Alg.~\ref{alg:global_pointer}. For simplicity, we omit the derivation from primal to QCQP and QCQP to SDP. For detailed derivation, please refer to the supplementary material.

\PAR{Pose-Only Optimization.}
\label{subsugbsection:Point-to-Plane Pose-only Optimization}
After fixing the plane parameters, we can readily convert the original plane adjustment problem in \cref{eq:point-to-plane PA} to a pose-only SDP as:
\begin{equation}
	\begin{aligned}
		\{\mathbf{X}^*\} = arg \min_{\mathbf{X}} \quad & \sum_{i=1}^{m} trace (  \mathcal{C}(i)  \mathbf{X}_i)\\
		\text{s.t. } \quad & \mathcal{C}(i) = \sum_{j = 1}^{n}(\mathbf{b}_j\mathbf{b}_j^{\top}) \otimes \mathcal{B}(i, j), \\
        & \mathbf{X}_i \succeq \mathbf{0}, \quad \{\text{redundant rotation constraints}\}, \\
		& i = 1, \ldots, m, \quad j = 1, \ldots, n,
	\end{aligned}
	\tag{Pose-Only SDP}
        \label{eq:pose_only_sdp}
\end{equation}
where the auxiliary vector $\mathbf{b}_j\in \mathbb{R}^4$ is defined as $\mathbf{b}_j = [\mathbf{n}_j ;-\mathbf{n}_j^{\top} \mathbf{q}_j]$ and rank one symmetric PSD matrix $\mathbf{X}_i = vec(\mathbf{T}_i) vec(\mathbf{T}_i)^{\top}$ represents the $i^{th}$ primal pose variable. The redundant rotation constraints are defined as:
\begin{align}
	\{\text{redundant rotation constraints}\} : \left\{\begin{matrix}
		&\mathbf{R}_i^{\top}\mathbf{R}_i = \mathbf{I}_3, \quad \mathbf{R}_i\mathbf{R}_i^{\top} = \mathbf{I}_3, \\
		&(\mathbf{R}_i\mathbf{e}_i) \times (\mathbf{R}_i\mathbf{e}_j) = (\mathbf{R}_i\mathbf{e}_k), \\
		&\forall (i, j, k) \in \left \{ (1, 2, 3), (2, 3, 1), (3, 1, 2)\right \}. 
	\end{matrix}\right.
\end{align}

\begin{algorithm}[t]
    \SetKwInOut{Input}{Input}
    \SetKwInOut{Output}{Output}
    \Input{Random initialized LiDAR poses $\{\mathbf{R}_i,\mathbf{t}_i\mid i=1,\dots,m\}$, random initialized plane parameters $\{\mathbf{n}_j,\mathbf{q}_j\mid j=1,\dots,n\}$, and labeled LiDAR points $\{\mathbf{P}_k\mid k\in obs(i,j)\}$.}
    \Output{Optimized LiDAR poses and plane parameters $\{\mathbf{R}^*,\mathbf{t}^*,\mathbf{n}^*,\mathbf{q}^*\}$.}
    Find the optimal local plane parameters $\{\mathbf{n}^{l*}, {d}^{l*}\}$\;
    Global normal direction calibration by solving the \ref{eq:rotation_only_sdp}\;
    \While{(not reach max iteration) and (not satisfy stopping criteria)}
    {
        Find LiDAR poses $\{\mathbf{R}^*,\mathbf{t}^*\}$ by solving \cref{eq:pose_only_closed}\;
        Find plane parameters $\{\mathbf{n}^*,\mathbf{q}^*\}$ by solving \cref{eq:plane_only_closed}\;
    }
    \caption{\textit{GlobalPointer++}}
    \label{alg:global_pointer++}
\end{algorithm}

\PAR{Plane-Only Optimization.}
\label{subsubsection:Point-to-Plane Plane-only Optimization}
Similar to pose-only optimization, we can conduct plane-only optimization by fixing the pose parameters. This reformulates the original plane adjustment problem in \cref{eq:point-to-plane PA} into a plane-only SDP as:
\begin{equation}
	\begin{aligned}
		\{\mathbf{Y}^{*}\} = arg \min_{\mathbf{Y}} \quad & \sum_{j=1}^{n}trace(\mathcal{D}(j)\mathbf{Y}_j)\\
		\text{s.t. } \quad
        & \mathcal{D}(j) = \sum_{i = 1}^{m}\mathbf{T}_i \mathcal{B}(i, j)\mathbf{T}_i^{\top}, \\
        & \mathbf{Y}_j \succeq \mathbf{0}, \quad \left \| \mathbf{n}_j \right \| = 1, \\
		& i = 1, \ldots, m, \quad j = 1, \ldots, n,
	\end{aligned}
	\tag{Plane-Only SDP}
        \label{eq:plane_only_sdp}
\end{equation}
where the rank one symmetric PSD matrix $\mathbf{Y}_j = \mathbf{b}_j\mathbf{b}_j^{\top}$ denotes the $j^{th}$ primal plane variable.

\subsection{GlobalPointer++}
\label{subsec:GlobalPointer++}
In this subsection, we introduce a new variant of the plane adjustment algorithm, \textit{GlobalPointer++}, which accelerates the original \textit{GlobalPointer} method. Similar to \textit{GlobalPointer}, \textit{GlobalPointer++} decouples the original formulation and alternately solves each sub-problem in closed form until convergence. Unlike \textit{GlobalPointer}, \textit{GlobalPointer++} relies on the plane-to-plane error defined in \cref{eq:plane-to-plane PA} for registration instead of the point-to-plane error. Although this formulation can be solved in closed form, the ambiguity of the normal direction prevents practical usage, as discussed in \cref{subsubsection:PA with Plane-to-Plane}. In the following, we first address the ambiguity issue by introducing global normal direction calibration and then derive the closed-form solvers. The algorithm is summarized in Alg.~\ref{alg:global_pointer++}.

\PAR{Global Normal Direction Calibration.}
To resolve the ambiguity of the normal direction, we reformulate the simultaneous rotation and normal direction search problem as a new SDP problem. Firstly, we rewrite the cross term $-2 \mathbf{n}_j^\top \mathbf{R}_i\mathbf{n}_{ij}^{l*}$  in ${\left \| \mathbf{n}_j - \mathbf{R}_i\mathbf{n}_{ij}^{l*} \right \|}_2^2$ (\cref{eq:plane-to-plane PA}) using quaternion-based rotation representation as $\mathbf{q}_i^{\top} \mathbf{M}_{ij} \mathbf{q}_i\boldsymbol{\theta}_{ij}$, where $\mathbf{q}_i$ is the unit quaternion of rotation $\mathbf{R}_i$, $\mathbf{M}_{ij}$ is the corresponding auxiliary matrix, and $\boldsymbol{\theta}_{ij}=\{+1,-1\}$ represents the normal direction sign. We then define a single column vector $\bar{\mathbf{q}}_i = [\mathbf{q}_i; \mathbf{q}_i \boldsymbol{\theta}_{i1}; \ldots; \mathbf{q}_i \boldsymbol{\theta}_{in}]$ and the corresponding rank one symmetric PSD matrix $\mathbf{Q}_i=\bar{\mathbf{q}}_i\bar{\mathbf{q}}_i^{\top}$. This can be relaxed to a rotation-only SDP as:
\begin{equation}
	\begin{aligned}
		\{\mathbf{Q}^{*}\} = arg \min_{\mathbf{Q}} \quad & \sum_{i = 1}^{m}  trace(\bar{\mathbf{M}}_{i} \mathbf{Q}_i) \\
		\text{s.t. } \quad 
        & \mathbf{Q}_i  \succeq \mathbf{0}, \quad trace([\mathbf{Q}_i]_{00}) = 1, \quad [\mathbf{Q}_i]_{00} = [\mathbf{Q}_i]_{jj}\\
		& i = 1, \ldots, m, \quad j = 1, \ldots, n,
	\end{aligned}
 \tag{Rotation-Only SDP}
 \label{eq:rotation_only_sdp}
\end{equation}
where $\bar{\mathbf{M}}_i$ includes all $\mathbf{M}_{ij}$ as
\begin{equation}
	\begin{aligned}
		[\bar{\mathbf{M}}_i]_{0j} = 0.5 \mathbf{M}_{ij}, \quad [\bar{\mathbf{M}}_i]_{j0} = 0.5 \mathbf{M}_{ij}^\top, \quad j = 1,\ldots, n.
	\end{aligned}
\end{equation}

{\it Remark.} Following this calibration, all local plane normal directions are made consistent. It is worth noting that this calibration can also be applied to two-frame registration. Consequently, we employ this solver to sequentially initialize the normal directions by incrementally registering point clouds, thereby reducing computational time. Once the normal directions are corrected, we then apply the following closed-form solvers to each sub-problem.

\PAR{Pose-Only Closed-Form Solver.}
After fixing the plane parameters, we can convert the original plane adjustment problem in \cref{eq:plane-to-plane PA} to a pose-only closed-form solver as:
\begin{equation}
	\begin{aligned}
		\{\mathbf{R}^{*} , \mathbf{t}^{*}\} = arg \min_{\mathbf{R}, \mathbf{t}} \quad & \sum_{i = 1}^{m} \sum_{j = 1}^{n} {\left \| \mathbf{n}_j - \mathbf{R}_i\mathbf{n}_{ij}^{l*} \right \|}_2^2 + {\left \| \mathbf{n}_{ij}^{l*\top}\mathbf{R}_i^{\top}\mathbf{t}_i  + d_j - {d}_{ij}^{l*}  \right \|}_2^2\\
		=arg \min_{\mathbf{q}, \mathbf{t}} \quad & \sum_{i = 1}^{m} \sum_{j = 1}^{n}\mathbf{q}_i^{\top} \mathbf{M}_{ij} \mathbf{q}_i+\left \| \mathbf{n}_{ij}^{l*\top}\mathbf{R}_i^{\top}\mathbf{t}_i  + d_j - {d}_{ij}^{l*}\right \|^2\\
		\text{s.t. } \quad
        & \left \| \mathbf{q}_i \right \| = 1, \quad i = 1, \ldots, m, 
	\end{aligned}
 \label{eq:pose_only_closed}
\end{equation}
where the auxiliary matrix $\mathbf{M}_{ij}$ is defined as in \ref{eq:rotation_only_sdp}. The quaternion can be analytically determined as the eigenvector corresponding to the minimum eigenvalue. Subsequently, by substituting the quaternion into the second term, the translation vector can be obtained in closed form using linear least squares.

\PAR{Plane-Only Closed-Form Solver.}
Similar to the pose-only closed-form solver, by fixing the plane parameters, we can convert the original plane adjustment problem in \cref{eq:plane-to-plane PA} to a plane-only closed-form solver as:
\begin{equation}
	\begin{aligned}
		\{\mathbf{n}^{*} , d^{*}\} = arg \min_{\mathbf{n}, d} \quad & \sum_{i = 1}^{m} \sum_{j = 1}^{n} {\left \| \mathbf{n}_j - \mathbf{R}_i\mathbf{n}_{ij}^{l*} \right \|}_2^2 + {\left \| \mathbf{n}_{ij}^{l*\top}\mathbf{R}_i^{\top}\mathbf{t}_i  + d_j - {d}_{ij}^{l*}  \right \|}_2^2\\
		\text{s.t. } \quad & \left \| \mathbf{n}_j \right \| = 1, \quad j = 1, \ldots, n.
	\end{aligned}
  \label{eq:plane_only_closed}
\end{equation}

The optimal $\mathbf{n}_j$ can be determined through eigen decomposition. Subsequently, the optimal $d_j$ can be analytically obtained by substituting the optimal $\mathbf{n}_j$.

\subsection{Performance Analysis}
\label{subsec:Performance Analysis}
In this section, we provide some intuitive remarks on our solver. For more detailed discussions, please refer to the supplementary material. 

{\it Remark 1.} For general scenes, the proposed \textit{Bi-Convex Relaxation} optimization strategy consistently ensures convergence to a local minimum of the original plane adjustment problem. Intuitively, the two sets of parameters are not mutually constrained by equalities or inequalities, making the overall problem more stable and straightforward.

{\it Remark 2.} In the absence of outliers and noise, each SDP sub-problem (\ie, \ref{eq:pose_only_sdp}, \ref{eq:plane_only_sdp}, and \ref{eq:rotation_only_sdp}) converges to a global minimum with zero duality gap. Moreover, even in the presence of low noise levels, our algorithm maintains empirical global optimality.

{\it Remark 3.} Although our original problem is non-convex, experimental results demonstrate that our algorithm achieves empirical global optimality. While this global optimality is empirical, further analysis can be conducted through other means, which is beyond the scope of this paper. 

{\it Remark 4.} The overlap between planes and poses is a critical factor for our algorithm. Problems with a larger overlap result in a more stable and efficient optimization convergence process. From a practical perspective, a larger overlap is preferred.

{\it Remark 5.} The time complexity of the proposed \textit{GlobalPointer} is linear with respect to the number of planes and poses. The time complexity of the proposed \textit{GlobalPointer++} is linear when the number of planes and poses is small, and quadratic when both are extremely large.
\begin{figure}[t]
	\centering
	\includegraphics[width=0.99\linewidth]{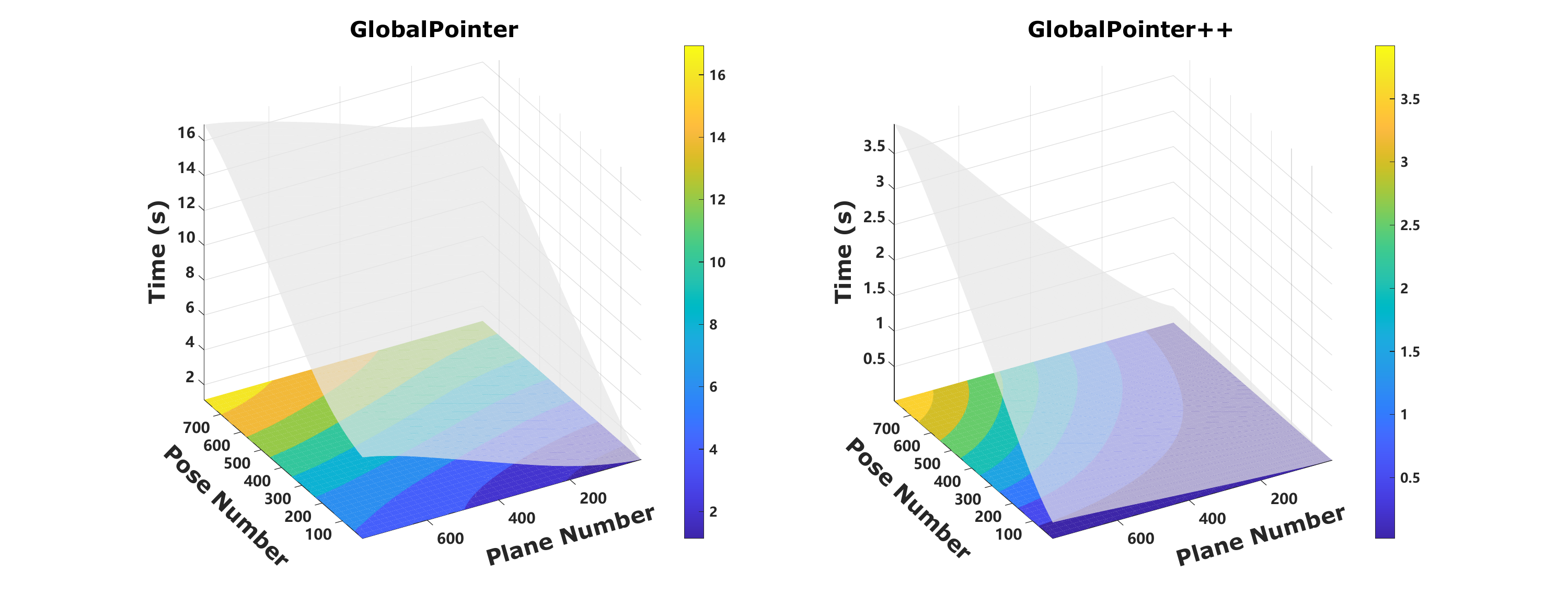}
	\caption{\textbf{Total optimization time analysis with increasing poses and planes.}}
	\label{fig:runtime}
\end{figure}

\section{Experiments}
\label{sec:Experiments}

\subsection{Testing Setup}
In our experiments, the proposed method is compared to the state-of-the-art methods including BALM2 \cite{BALM2}, EF \cite{EF}, ESO-Full \cite{ESO}, ESO-BFGS (the first-order version of \cite{ESO}), PA-Full \cite{zhou2020efficient}, and PA-Decoupled (the decoupled version of \cite{zhou2020efficient}). We implement our method in Matlab and run it on a laptop with an i9-13900HX CPU and 32 GB RAM. The maximum number of iterations is set to 200 for all second-order solvers, and the relative stop tolerance is set to $10^{-4}$. We use Yalmip as our solver's interface and Mosek as the core SDP solver.

\begin{figure}[t]
	\centering
	\includegraphics[width=0.96\linewidth]{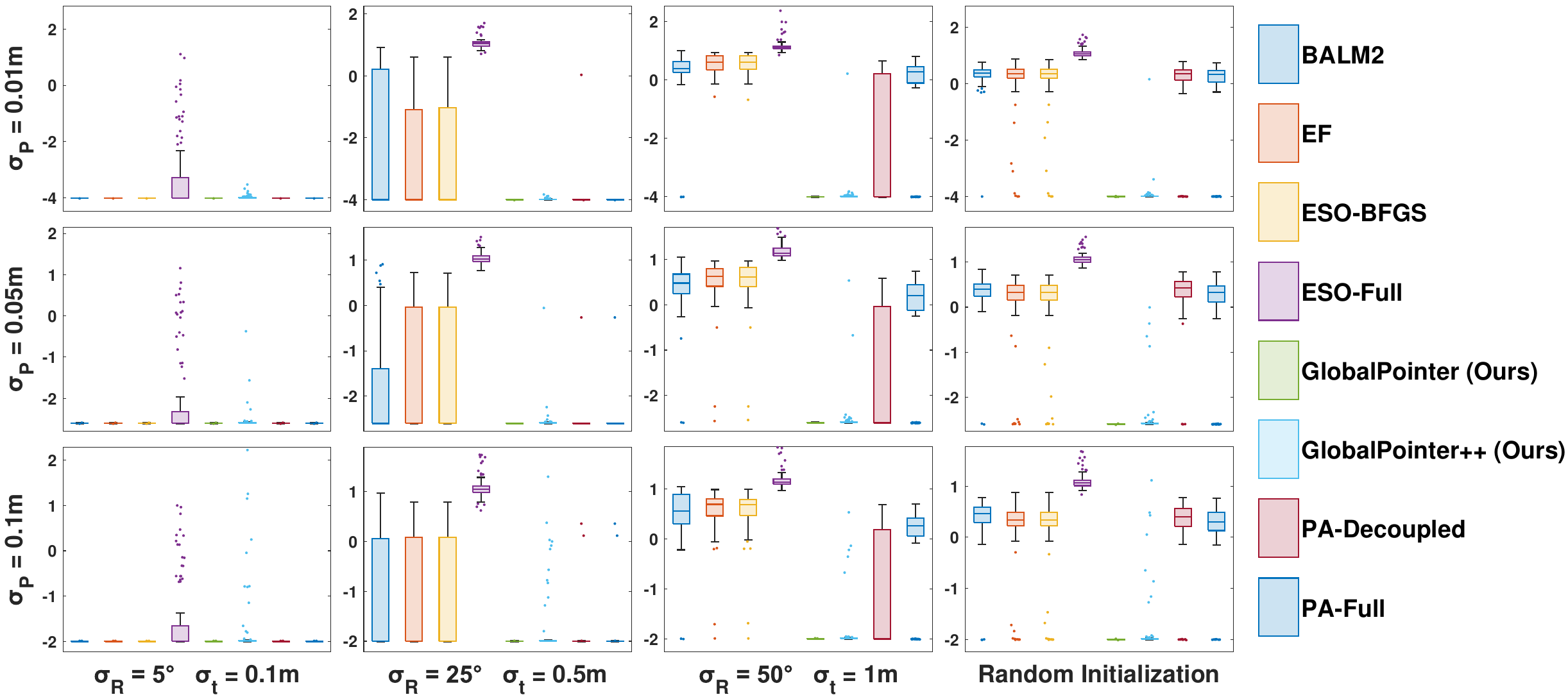}
	\caption{\textbf{Accuracy comparisons on the synthetic dataset under varying point cloud noise levels and pose initialization noise levels.} The $y$ axis represents the total point-to-plane error $e_{total}$ in the log10 scale.}
	\label{fig:profile}
\end{figure}

\subsection{Synthetic Data}
\PAR{Testing Setup.} We generate numerous virtual planes and virtual observation poses. Each pose is set to be inside a box with a maximum size of 50 meters, and each plane is set to be observed from any pose.
\begin{figure}[t]
	\centering
\includegraphics[width=0.99\linewidth]{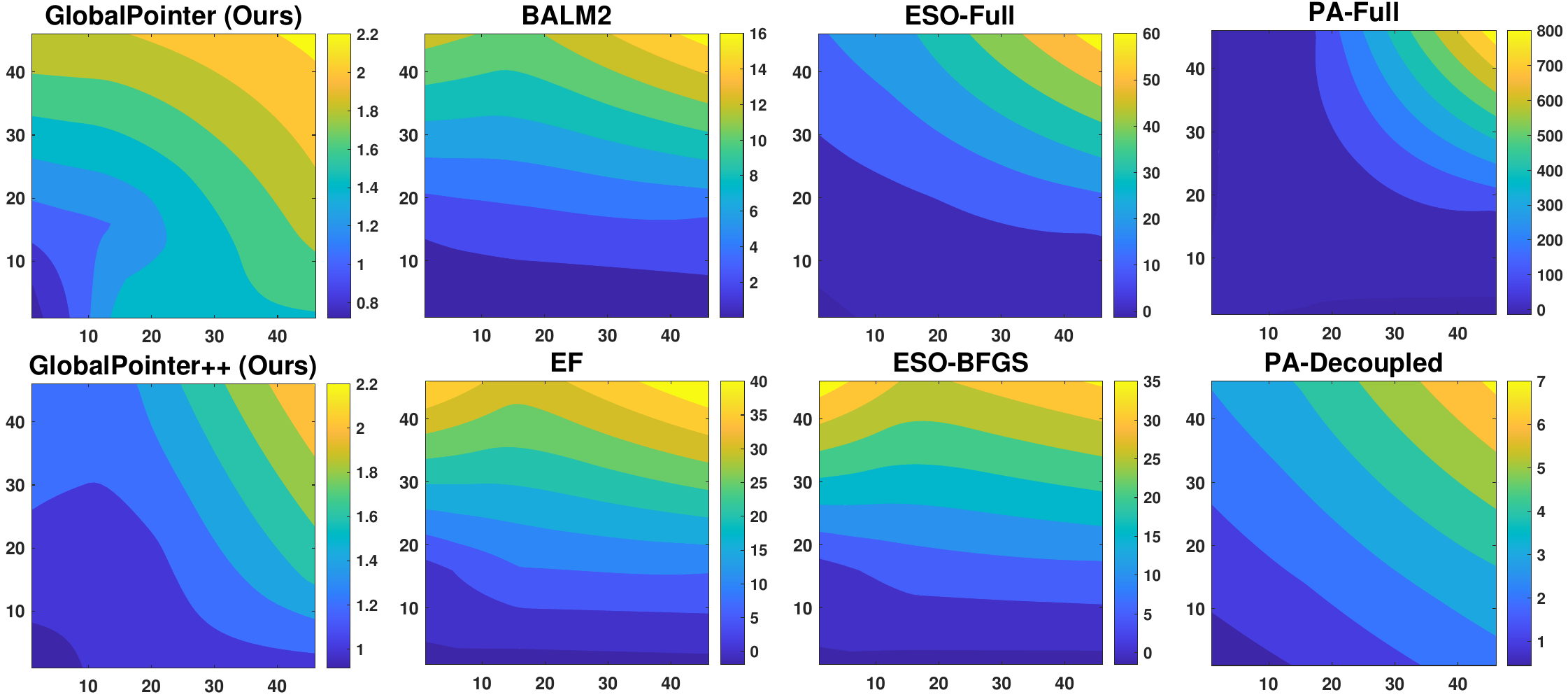}
	\caption{\textbf{Relative time complexity comparisons.} The $x$ axis represents the number of planes and the $y$ axis represents the number of poses. The color bar on the right side of each subplot indicates the mapping relationship between the multiplier of runtime growth and the colors.}
	\label{fig:time complexity}
\end{figure}

\PAR{Runtime.}
We evaluate the runtime of {\it GlobalPointer} and {\it GlobalPointer++}. The number of planes and poses increases gradually, using well-initialized parameters. All statistics are computed over 50 independent trials. As illustrated in \cref{fig:runtime}, the runtime of {\it GlobalPointer} exhibits a linear growth trend with increasing numbers of planes and poses, while {\it GlobalPointer++} achieves a 5-10x speedup compared to {\it GlobalPointer}. Interestingly, as the number of planes and poses reaches a certain magnitude, the most time-consuming part in the {\it GlobalPointer++} is neither the eigen decomposition nor the linear least squares, but rather the data preparation for each iteration. Due to the quadratic growth complexity of data preparation time, the entire algorithm demonstrates quadratic time complexity when the number of planes and poses is large.

\PAR{Accuracy.}
We extensively compare our proposed algorithms with other methods in terms of accuracy under varying levels of point cloud noise and pose initialization noise. We choose the total point-to-plane error, defined as $e_{total} = (\mathbf{n}^{\top}(\mathbf{R}\mathbf{P}+\mathbf{t}) + d)^2$, as the evaluation metric. All statistics are computed over 50 independent trials. As shown in \cref{fig:profile}, {\it GlobalPointer} consistently converges to the global optimum across all settings, while {\it GlobalPointer++} is more sensitive to point cloud noise. ESO \cite{ESO} performs poorly due to unstable numerical stability in Hessian matrix derivation. Other methods fail to converge to the global minimum above a certain level of noise.

\PAR{Time Complexity.}
We also evaluate the growth of time complexity for the algorithms. Starting with 5 planes and 5 poses, we incrementally increase the number of poses and planes. Time complexity is measured as the multiple of the optimization time relative to the initial setting. We conduct 50 independent trials and use the median time as the metric. Experimental results in \cref{fig:time complexity} demonstrate that our two algorithms exhibit similar time complexity. BALM2 \cite{BALM2}, EF \cite{EF}, and ESO-BFGS \cite{ESO} show less sensitivity to increasing numbers of planes. However, ESO-Full \cite{ESO} and PA-Full \cite{zhou2020efficient} exhibit time complexities approaching cubic. PA-Decoupled  \cite{zhou2020efficient}, which also employs alternating minimization, demonstrates time complexity similar to our method.

\subsection{Real Data}
\begin{figure}[t]
	\centering
	\includegraphics[width=0.99\linewidth]{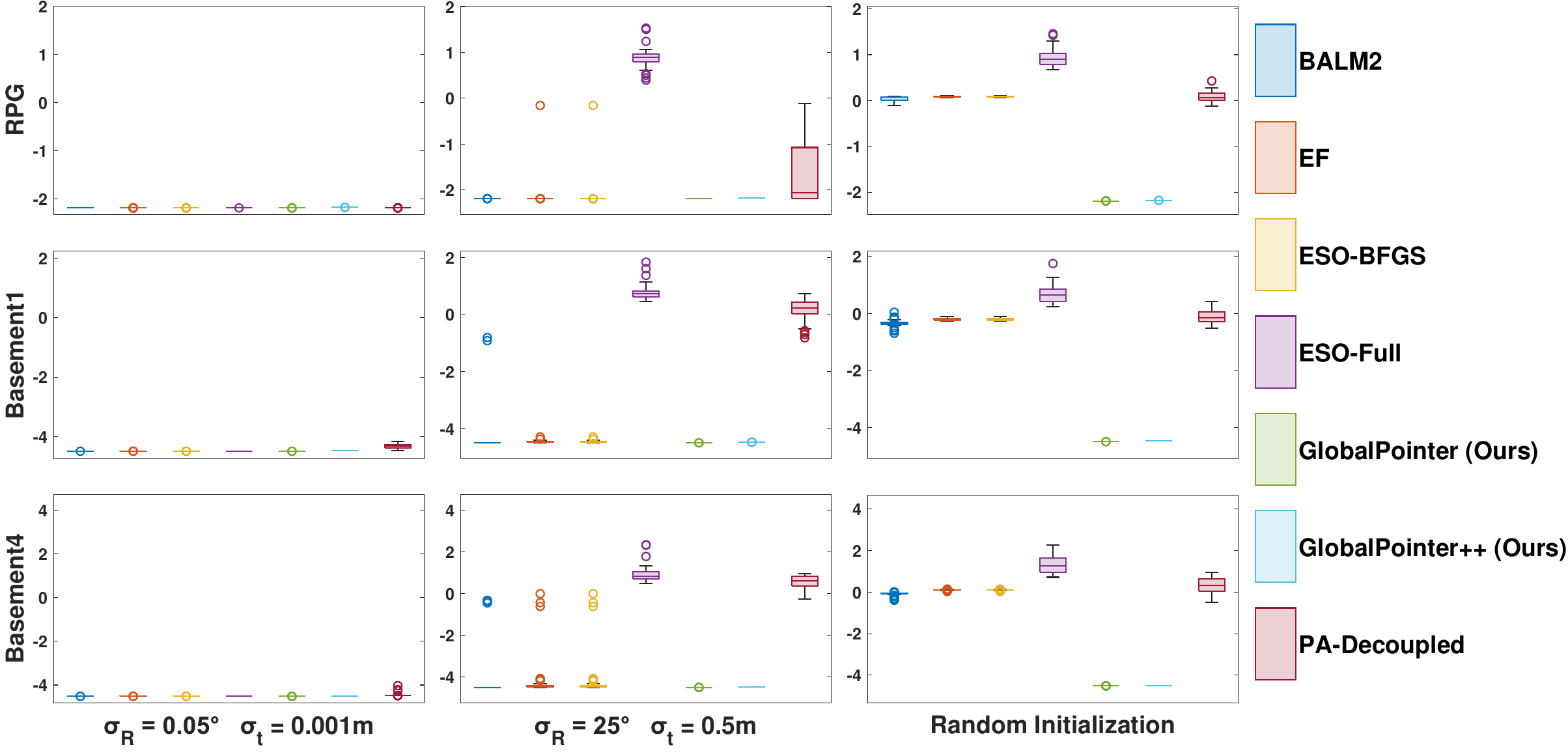}
	\caption{\textbf{Accuracy comparisons on the real dataset under varying pose initialization noise levels.} The $y$ axis represents the total point-to-plane error $e_{total}$ in the log10 scale.}
	\label{fig:real profile}
\end{figure}

\PAR{Accuracy.}
For real datasets, we select sequences from the Hilti dataset \cite{helmberger2022hilti} with a higher number of indoor planes. Initially, we transform the local point clouds of each frame into the global coordinate system using ground truth poses. Subsequently, we employ plane segmentation and fitting algorithms to extract point clouds  \cite{zhou2018open3d} associated with the same plane, with a RANSAC fitting threshold of $0.01$. The segmentation results can be found in
the supplementary material. Furthermore, to align with our hypothesis that each plane should be observed by as many LiDAR frames as possible, we randomly divide all point cloud frames into 50 subsets and accumulate them into individual sub-cloud frames. All statistics are computed over 50 independent trials. In \cref{fig:real profile}, we compare the accuracy of our algorithms with other methods. It is evident that our proposed {\it GlobalPointer} and {\it GlobalPointer++} consistently demonstrate global convergence performance across all settings, while other methods require well-initialized values to converge to the ground truth.
\section{Conclusions}
In this paper, we exploit a novel optimization strategy, \textit{Bi-Convex Relaxation}, and apply it to solve the plane adjustment problem with two variants of errors, resulting in the proposals of \textit{GlobalPointer} and \textit{GlobalPointer++}, respectively. Extensive synthetic and real experiments demonstrate that our method enables large-scale plane adjustment with linear time complexity, larger convergence region, and without relying on good initialization, while achieving similar accuracy as prior methods.

\section*{Acknowledgments}
This work was supported in part by NSFC under Grant 62202389, in part by a grant from the Westlake University-Muyuan Joint Research Institute, and in part by the Westlake Education Foundation

\appendix
\section*{SUPPLEMENTARY MATERIAL}
\section*{Overview}
In this supplementary material, we further provide the following content:
\begin{itemize}
    \item A complete derivation of \textit{GlobalPointer} (\cref{supp:sec:Supplementary Derivation}).
    \item Detailed discussions on the remarks presented in in \cref{subsec:Performance Analysis} of the main paper (\cref{supp:sec:Supplementary Proof}).
    \item Additional experiment analysis of our solvers (\cref{supp:sec:Supplementary Experiment}).
\end{itemize}

\section{Supplementary Derivation}
\label{supp:sec:Supplementary Derivation}

In this section, we derive the primal QCQP and the corresponding SDP with convex relaxation for each sub-problem in \textit{GlobalPointer}.

\PAR{Pose-Only Optimization.}
\label{subsugbsection:Point-to-Plane Pose-only Optimization}
After fixing the plane parameters, we can readily convert the original plane adjustment problem defined in Eq.~3 to a pose-only QCQP as:

\begin{equation}
	\begin{aligned}
		(vec(\mathbf{T}^{*})) = arg \min_{vec(\mathbf{T})} \quad & \sum_{j = 1}^{n} \sum_{i = 1}^{m} \mathbf{b}_j^{\top}\mathbf{T}_i \mathcal{B}(i, j)\mathbf{T}_i^{\top}\mathbf{b}_j\\
		= arg \min_{vec(\mathbf{T})} \quad & \sum_{i = 1}^{m} vec(\mathbf{T}_i)^{\top} \mathcal{C}(i)  vec(\mathbf{T}_i)\\
		\text{s.t. } \quad & \mathcal{C}(i) = \sum_{j = 1}^{n}(\mathbf{b}_j\mathbf{b}_j^{\top}) \otimes \mathcal{B}(i, j), \\
        & \mathbf{R}_i \in  \mathbf{SO(3)}, \\
		& i = 1, \ldots, m, \quad j = 1, \ldots, n,
	\end{aligned}
	\tag{Pose-Only QCQP}
\end{equation}
where the auxiliary vector $\mathbf{b}_j$ is defined as $\mathbf{b}_j = [\mathbf{n}_j; -\mathbf{n}_j^{\top} \mathbf{q}_j]$.
This pose-only QCQP is a standard block-wise QCQP and can be decoupled into $m$ sub-problems, each of which can be converted into $m$ low-dimensional SDPs.  
Let the rank one symmetric PSD matrix $\mathbf{X}_i = vec(\mathbf{T}_i) vec(\mathbf{T}_i)^{\top}$ represent the $i^{th}$ primal pose variable. We can convert this pose-only QCQP to a pose-only SDP as:
\begin{equation}
	\begin{aligned}
		\{\mathbf{X}^*\} = arg \min_{\mathbf{X}} \quad & \sum_{i=1}^{m} trace (  \mathcal{C}(i)  \mathbf{X}_i)\\
		\text{s.t. } \quad & \mathcal{C}(i) = \sum_{j = 1}^{n}(\mathbf{b}_j\mathbf{b}_j^{\top}) \otimes \mathcal{B}(i, j), \\
        & \mathbf{X}_i \succeq \mathbf{0}, \quad \{\text{redundant rotation constraints}\}, \\
		& i = 1, \ldots, m, \quad j = 1, \ldots, n,
	\end{aligned}
	\tag{Pose-Only SDP}
        \label{eq:pose_only_sdp}
\end{equation}
where the redundant rotation constraints are defined as:
\begin{align}
	\{\text{redundant rotation constraints}\} : \left\{\begin{matrix}
		&\mathbf{R}_i^{\top}\mathbf{R}_i = \mathbf{I}_3, \quad \mathbf{R}_i\mathbf{R}_i^{\top} = \mathbf{I}_3, \\
		&(\mathbf{R}_i\mathbf{e}_i) \times (\mathbf{R}_i\mathbf{e}_j) = (\mathbf{R}_i\mathbf{e}_k), \\
		&\forall (i, j, k) \in \left \{ (1, 2, 3), (2, 3, 1), (3, 1, 2)\right \}. 
	\end{matrix}\right.
\end{align}

\PAR{Plane-Only Optimization}
\label{subsubsection:Point-to-Plane Plane-only Optimization}
Similar to pose-only optimization, we can conduct plane-only optimization by fixing the pose parameters. We can readily convert the original plane adjustment problem in Eq.~3 to a plane-only QCQP as:
\begin{equation}
	\begin{aligned}
		(\mathbf{b}^{*}) = arg \min_{\mathbf{b}} \quad & \sum_{j = 1}^{n} \sum_{i = 1}^{m} \mathbf{b}_j^{\top}\mathbf{T}_i \mathcal{B}(i, j)\mathbf{T}_i^{\top}\mathbf{b}_j \\
		= arg \min_{\mathbf{b}} \quad & \sum_{j = 1}^{n} \mathbf{b}_j^{\top}(\sum_{i = 1}^{m}\mathbf{T}_i \mathcal{B}(i, j)\mathbf{T}_i^{\top})\mathbf{b}_j \\
		\text{s.t. } \quad & \left \| \mathbf{n}_j \right \| = 1, \\ & i = 1, \ldots, m, \quad j = 1, \ldots, n.
	\end{aligned}
	\tag{Plane-Only QCQP}
\end{equation}

Similar to pose-only SDP, we introduce the rank one symmetric PSD matrix $\mathbf{Y}_j = \mathbf{b}_j\mathbf{b}_j^{\top}$ and convert this plane-only QCQP to a plane-only SDP as:
\begin{equation}
	\begin{aligned}
		\{\mathbf{Y}^{*}\} = arg \min_{\mathbf{Y}} \quad & \sum_{j=1}^{n}trace(\mathcal{D}(j)\mathbf{Y}_j)\\
		\text{s.t. } \quad & \mathcal{D}(j) = \sum_{i = 1}^{m}\mathbf{T}_i \mathcal{B}(i, j)\mathbf{T}_i^{\top} \\ & \mathbf{Y}_j \succeq \mathbf{0}, \quad \left \| \mathbf{n}_j \right \| = 1, \\
		& i = 1, \ldots, m, \quad j = 1, \ldots, n.
	\end{aligned}
	\tag{Plane-Only SDP}
        \label{eq:plane_only_sdp}
\end{equation}

\section{Detailed Discussions}
\label{supp:sec:Supplementary Proof}
In this section, we further discuss all the remarks claimed in \cref{subsec:Performance Analysis} of the main paper. 

\subsection{Remark 1}
{\it Remark 1. For general scenes, the proposed \textit{Bi-Convex Relaxation} optimization strategy consistently ensures convergence to a local minimum of the original plane adjustment problem. Intuitively, the two sets of parameters are not mutually constrained by equalities or inequalities, making the overall problem more stable and straightforward.}

Considering a multi-variable constrained optimization problem defined as:
\begin{equation}
	\begin{aligned}
		(\mathbf{a}^{*}, \mathbf{b}^{*}) = arg \min_{\mathbf{a}, \mathbf{b}} \quad & f(\mathbf{a}, \mathbf{b})\\
		\text{s.t. } \quad & c_i(\mathbf{a}, \mathbf{b}) \le  0, \quad i = 1,\ldots,m \\
		& c_j(\mathbf{a}, \mathbf{b}) = 0, \quad j = 1,\ldots,n.
	\end{aligned}
	\tag{Primal Constrained Problem}
        \label{supp:Primal Constrained Problem}
\end{equation}

By fixing one variable, we can solve the remaining sub-problem until convergence. The two sub-convex problems at iteration $n$ are defined as:
\begin{equation}
    \begin{aligned}
		(\mathbf{a}^{*}) = arg \min_{\mathbf{a}} \quad & f(\mathbf{a}, \mathbf{b})\\
		\text{s.t. } \quad & c_i(\mathbf{a}, \mathbf{b}) \le 0, \quad i = 1, \ldots, m, \\
		& c_j(\mathbf{a}, \mathbf{b}) = 0, \quad j = 1, \ldots, n, \\
		(\mathbf{b}^{*}) = arg \min_{\mathbf{b}} \quad & f(\mathbf{a}, \mathbf{b})\\
		\text{s.t. } \quad & c_i(\mathbf{a}, \mathbf{b}) \le 0, \quad i = 1, \ldots, m, \\
		& c_j(\mathbf{a}, \mathbf{b}) = 0, \quad j = 1, \ldots, n. \\
    \end{aligned}
    \tag{Bi-convex Problem}
    \label{supp:Bi-convex Problem}
\end{equation}

~\\
{\bf Lemma 1.} {\it The partial optimum \cite{gorski2007biconvex} of \ref{supp:Bi-convex Problem} is a local minimum of \ref{supp:Primal Constrained Problem} if the partial derivatives $c_i(\mathbf{a}, \mathbf{b})$ and $c_j(\mathbf{a}, \mathbf{b})$ with respect to $\mathbf{a}$ and $\mathbf{b}$ are linearly independent.}

~\\
{\it Proof.} Due to the lack of constraint correlation between $\mathbf{a}$ and $\mathbf{b}$ in our optimization problem, this sufficient condition can be easily verified as:
\begin{equation}
    \begin{aligned}
        &\frac{\partial c_i(a,b)}{\partial a} \text{ linearly independent of } \frac{\partial c_i(a,b)}{\partial b}\\
        &\frac{\partial c_j(a,b)}{\partial a} \text{ linearly independent of } \frac{\partial c_j(a,b)}{\partial b}
    \end{aligned}
\end{equation}

Therefore, for each sub-problem, we can further simplify it as:
\begin{equation}
    \begin{aligned}
		(\mathbf{a}^{*}) = arg \min_{\mathbf{a}} \quad & f(\mathbf{a}, \mathbf{b})\\
		\text{s.t. } \quad & c_i(\mathbf{a}) \le  0, \quad i = 1, \ldots, m, \\
		& c_j(\mathbf{a}) = 0, \quad j = 1, \ldots, n, \\
		(\mathbf{b}^{*}) = arg \min_{\mathbf{b}} \quad & f(\mathbf{a}, \mathbf{b})\\
		\text{s.t. } \quad & c_i(\mathbf{b}) \le  0, \quad i = 1, \ldots, m, \\
		& c_j(\mathbf{b}) = 0, \quad j = 1, \ldots, n. \\
    \end{aligned}
    \tag{Simplified Bi-convex Problem}
    \label{supp:Sim Bi-convex Problem}
\end{equation}

Once we achieve the partial optimum (where each sub-problem solver reaches a local minimum), we can combine the Karush-Kuhn-Tucker (KKT) conditions \cite{Boyd2010ConvexO} from each sub-problem into the larger KKT conditions. These larger KKT conditions are exactly the KKT conditions of the original problem.$\hfill\square$
~\\

Hence, once the bi-convex optimization problem converges to the partial optimum, we conclude that this solution converges to a local minimum of the primal constrained problem.

\subsection{Remark 2}
{\it Remark 2. In the absence of outliers and noise, each SDP sub-problem (\ie, \ref{eq:pose_only_sdp}, \ref{eq:plane_only_sdp}, and \ref{eq:rotation_only_sdp}) converges to a global minimum with zero duality gap. Moreover, even in the presence of a low noise level, our algorithm maintains empirical global optimality.}

To begin with, we define the primal QCQP as:
\begin{equation}
    \begin{aligned}
        \min_{\mathbf{x} \in \mathbb{R}^n} \quad & \mathbf{x}^{\top} \mathbf{C} \mathbf{x}\\
         \text{s.t. } \quad & \mathbf{x}^{\top} \mathbf{A}_i \mathbf{x} = b_i, \quad i = 1, \ldots, m.
    \end{aligned}
    \tag{Primal QCQP}
    \label{supp:eq:Primal QCQP}
\end{equation}

The Lagrange dual of the primal QCQP can be derived as:
\begin{equation}
    \begin{aligned}
        \max_{\mathbf{y} \in \mathbb{R}^m} \quad & \mathbf{b}^{\top}\mathbf{y}\\
         \text{s.t. } \quad & \mathbf{C} - \sum_{i = 1}^{m} y_i\mathbf{A}_i  \succeq 0.
    \end{aligned}
    \tag{Dual SDP}
    \label{supp:eq:Dual SDP}
\end{equation}

The dual of the Lagrange dual of the primal QCQP can be derived as:
\begin{equation}
    \begin{aligned}
        \min_{\mathbf{X} \in \mathcal{S}^n} \quad & trace(\mathbf{C}\mathbf{X})\\
         \text{s.t. } \quad & trace(\mathbf{A}_i\mathbf{X}) = b_i, \quad i = 1, \ldots, m, \\
         & \mathbf{X} \succeq \mathbf{0}.
    \end{aligned}
    \tag{Dual Dual SDP}
    \label{supp:eq:Dual Dual SDP}
\end{equation}

Based on Lemma 2.1 in \cite{cifuentes2022local}, we define the following lemmas: 

~\\
{\bf Lemma 2.} {\it Let $\mathcal{H}(\mathbf{y})=\mathbf{C} - \sum_{i = 1}^{m} y_i\mathbf{A}_i$. $\mathbf{x}$ is proven to be optimal for \ref{supp:eq:Primal QCQP}, and strong duality holds between \ref{supp:eq:Primal QCQP} and \ref{supp:eq:Dual SDP} if $\mathbf{x} \in \mathbb{R}^{n}, \mathbf{y} \in \mathbb{R}^m$ satisfy:
\begin{equation}
    \left\{\begin{matrix}
    \mathbf{x}^{\top} \mathbf{A}_i \mathbf{x} = b_i, \quad i = 1, \ldots, m &(\text{Primal Feasibility})\\
    \mathcal{H}(\mathbf{y}) \succeq 0 &(\text{Dual Feasibility})\\
    \mathcal{H}(\mathbf{y})\mathbf{x} = 0 &(\text{Stationary Condition})
    \end{matrix}\right.
\end{equation}
}

~\\
{\bf Lemma 3.} {\it In addition to Lemma 2, if $\mathcal{H}(\mathbf{y})$ has corank one, then $\mathbf{x}\mathbf{x}^{\top}$ is the unique optimum of \ref{supp:eq:Dual Dual SDP}, and $\mathbf{x}$ is the unique optimum of \ref{supp:eq:Dual SDP}.}
~\\

Based on these two lemmas, we can prove the strong duality of our methods.

~\\
{\bf Theorem 1.} {\it The duality gap of Pose-Only SDP is zero under the noise-free and outlier-free condition.}

~\\
{\it Proof.} Let $\mathbf{x}_i^{*}$ be the ground truth solution, and let the corresponding Lagrange multipliers $\mathbf{y}_i$ reduce to 0. Then $\mathcal{H}_i(\mathbf{y}_i) = \mathcal{C}(i)$. The primal feasibility is always satisfied. For the dual feasibility, it is straightforward to show that:
\begin{equation}
\begin{aligned}
        0 = \mathbf{x}_i^{*\top}\mathcal{H}_i(\mathbf{y}_i)\mathbf{x}_i^{*} \le  \mathbf{x}_i^{\top}\mathcal{H}_i(\mathbf{y}_i)\mathbf{x}_i\\ \forall  \mathbf{x}_i \in \{\mathbf{x}|\mathbf{x}^{\top} \mathbf{A}_i \mathbf{x} = b_i\}.
\end{aligned}
\end{equation}
which guarantees the dual feasibility condition. For the stationary condition, we can perform Cholesky decomposition as $\mathcal{H}_i(\mathbf{y}_i) = \mathbf{L}_i \mathbf{L}_i^{\top}$. Since $\mathbf{L}_i^{\top}\mathbf{x}_i=0$, then  $\mathcal{H}_i(\mathbf{y}_i)\mathbf{x}_i = \mathbf{L}_i \mathbf{L}^{\top}_i\mathbf{x}_i = 0$, satisfying the stationary condition. Finally, given the plane correspondence, the ground truth rigid transformation $\mathbf{x}^{*}_i\mathbf{x}^{\top*}_i$ is the unique non-zero solution that satisfies $trace(\mathcal{H}_i(\mathbf{y}_i)\mathbf{x}_i\mathbf{x}_i^{\top}) = 0$ up to scale. We can conclude that $\mathcal{H}_i(\mathbf{y}_i)$ is corank one, and thus proving Theorem 1. $\hfill\square$

~\\
{\bf Theorem 2.} {\it The duality gap of Plane-Only SDP is zero under the noise-free and outlier-free condition.}

~\\
{\it Proof.} Let $\mathbf{b}_j^{*}$ be the ground truth solution, and let the corresponding Lagrange multipliers $\mathbf{y}_j$ reduce to 0. Then $\mathcal{H}_j(\mathbf{y}_j) = \mathcal{D}(j)$. The primal feasibility is always satisfied. For the dual feasibility, it is straightforward to show that:
\begin{equation}
\begin{aligned}
        0 = \mathbf{b}^{*\top}_j\mathcal{H}_j(\mathbf{y}_j) \mathbf{b}^{*}_j \le  \mathbf{b}^{\top}_j\mathcal{H}_j(\mathbf{y}_j)\mathbf{b}_j\\ \forall  \mathbf{b}_j \in \{\mathbf{x}|\mathbf{x}^{\top} \mathbf{A}_i \mathbf{x} = b_i\}.
\end{aligned}
\end{equation}
which guarantees the dual feasibility condition. For the stationary condition, we can perform Cholesky decomposition as $\mathcal{H}_j(\mathbf{y}_j) = \mathbf{L}_j \mathbf{L}_j^{\top}$. As $\mathbf{L}_j^{\top}\mathbf{b}_j=0$, then  $\mathcal{H}_j(\mathbf{y}_j)\mathbf{b}_j = \mathbf{L}_j \mathbf{L}^{\top}_j\mathbf{b}_j = 0$, satisfying the stationary condition. Finally, given the pose correspondence, the ground truth plane $\mathbf{b}^{*}_j\mathbf{b}^{\top*}_j$ is the unique non-zero solution that satisfies $trace(\mathcal{H}_j(\mathbf{y}_j)\mathbf{b}_j\mathbf{b}_j^{\top}) = 0$ up to scale. We can conclude that $\mathcal{H}_j(\mathbf{y}_j)$ is corank one, thus proving Theorem 2.  $\hfill\square$

~\\
{\bf Theorem 3.} {\it The duality gap of Rotation-Only SDP is zero under the noise-free and outlier-free condition.}

~\\
{\it Proof.} Let $\mathbf{q}^{*}$ be the ground truth solution, and set the corresponding Lagrange multipliers $\mathbf{y}$ to -2. Then $\mathcal{H}(\mathbf{y}) = \mathbf{M} + 2\mathbf{I}$. The primal feasibility is always satisfied. For the dual feasibility, it is straightforward to show that:
\begin{equation}
\begin{aligned}
        0 = \mathbf{q}^{*\top}(\mathbf{M} + 2\mathbf{I})\mathbf{q}^{*} \le  \mathbf{x}^{\top}(\mathbf{M} + 2\mathbf{I})\mathbf{x}\\
        \forall  \mathbf{x} \in \{\mathbf{x}|\mathbf{x}^{\top} \mathbf{A}_i \mathbf{x} = b_i\}
\end{aligned}
\end{equation}
which also verifies the stationary condition. As $\mathbf{q}^{*}$ denotes the unique non-zero solution of $\mathbf{q}^{\top}(\mathbf{M} + 2\mathbf{I})\mathbf{q} = 0$ up to scale, we can conclude that $\mathcal{H}(\mathbf{y})$ is corank one, thus proving Theorem 3. $\hfill\square$

\subsection{Remark 3}
{\it Remark 3. Although our original problem is non-convex, experimental results demonstrate that our algorithm achieves empirical global optimality. While this global optimality is empirical, further analysis can be conducted through other means, which is beyond the scope of this paper.} 

In general, when the original problem is convex, utilizing the alternating minimization ensures global convergence. However, for the plane adjustment problem, the non-convex nature of the original problem prevents our algorithms from guaranteeing convergence to the global minimum. Nevertheless, our proposed algorithms exhibit several advantages: 
\begin{itemize}
    \item By transforming the sub-problem into a convex SDP problem, the convergence region is significantly enlarged, especially when utilizing alternating minimization. 
    \item In specific scenarios, particularly under fully observed scenarios where each plane is observed by every LiDAR frame, our algorithm demonstrates global optimality, empirically validated in \cref{supp:subsec:Global Optimality Check}.
\end{itemize}

\subsection{Remark 4}
{\it Remark 4. The overlap between planes and poses is a critical factor for our algorithm. Problems with a larger overlap result in a more stable and efficient optimization convergence process. From a practical perspective, a larger overlap is preferred.}

As shown in \cref{supp:subsec:Overlap Analysis}, the computational time increases from 3s to 10s as the overlap ratio decreases from 100\% to 20\% while achieving consistent accuracy. This confirms our remark that a small overlap ratio may cause vibration in the convergence process.

\subsection{Remark 5}
{\it Remark 5. The time complexity of the proposed GlobalPointer is linear with respect to the number of planes and poses. The time complexity of the proposed GlobalPointer++ is linear when the number of planes and poses is small, and quadratic when both are extremely large.}

For \textit{GlobalPointer}, assuming a total of $I$ iterations with $m$ poses and $n$ planes, where the optimization time of each single SDP is fixed, the total SDP optimization time complexity for pose-only and plane-only optimization is $\mathcal{O}(Im)$ and $\mathcal{O}(In)$, respectively. The accumulation of $\mathcal{B}(i, j)$ is performed once before optimization, which can be considered negligible. Assuming a fixed calculation time for $(\mathbf{b}_j\mathbf{b}_j^{\top}) \otimes \mathcal{B}(i, j)$ and $\mathbf{T}_i \mathcal{B}(i, j)\mathbf{T}_i^{\top}$, the time complexity for data preparation is $\mathcal{O}(Imn)$. Furthermore, since the SDP optimization time is several orders of magnitude longer than the data preparation time, the total computational time complexity is $\mathcal{O}(Imn+Im+In)\approx \mathcal{O}(Im+In)$. Thus, we conclude that the time complexity of \textit{GlobalPointer} is linear with respect to the number of poses and planes.

For \textit{GlpbalPointer++}, assuming a total of $I$ iterations with $m$ poses and $n$ planes, the time complexity for global normal direction calibration is $\mathcal{O}(mn^2)$. The pose-only and plane-only closed-form solvers have fixed calculation time per pose or plane, resulting in time complexities of $\mathcal{O}(Im)$ and $\mathcal{O}(In)$, respectively. Similar to \textit{GlobalPointer}, the time complexity for data preparation remains $\mathcal{O}(Imn)$. Thus, the total computational time complexity is $\mathcal{O}(Imn+Im+In)$. When $mn$ is not excessively large, the dominant time complexity of \textit{GlobalPointer++} is linear with respect to the number of poses and planes. However, as $mn$ increases, the quadratic increase in data preparation time makes \textit{GlobalPointer++} quadratic with respect to the number of poses and planes. Despite this quadratic complexity, the absolute computation time does not become significantly large.

\section{Supplementary Experiment}
\label{supp:sec:Supplementary Experiment}
In this section, we present additional experimental results on Global Optimality Analysis~\ref{supp:subsec:Global Optimality Check}, Synthetic Data Analysis~\ref{supp:subsec:Synthetic Data Analysis}, Real Data Analysis~\ref{supp:subsec:Real Data Analysis}, Absolute Time Analysis~\ref{supp:subsec:Absolute Time Analysis}, and Overlap Analysis~\ref{supp:subsec:Overlap Analysis}.
\subsection{Evaluation Metrics}
Given the estimated rotation $\mathbf{R}$, translation $\mathbf{t}$, normal vector $\mathbf{n}$, and $d$ (where $d=-\mathbf{n}^{\top}\mathbf{q}$ as introduced in the point-to-plane error), we define the following metrics as:
\begin{equation}
\left\{
\begin{array}{rcl}
e_{total} = (\mathbf{n}^{\top}(\mathbf{R}\mathbf{P}+\mathbf{t}) + d)^2\\
e_{R} = arccos(({trace(\mathbf{R}_{g}^{\top}\mathbf{R})-1})/{2})^2\\
e_{t} = ||\mathbf{t}-\mathbf{t}_g||^2_2\\
e_{n} = ||abs(\mathbf{n})-abs(\mathbf{n}_g)||^2_2\\
e_{d} = ||abs(d)-abs(d_g)||^2_2
\end{array} \right. 
\end{equation}
where $[.]_g$ denotes the ground truth. 

\subsection{Global Optimality Analysis}
\label{supp:subsec:Global Optimality Check}

\begin{figure}[ht]
	\centering
	\includegraphics[width=0.95\linewidth]{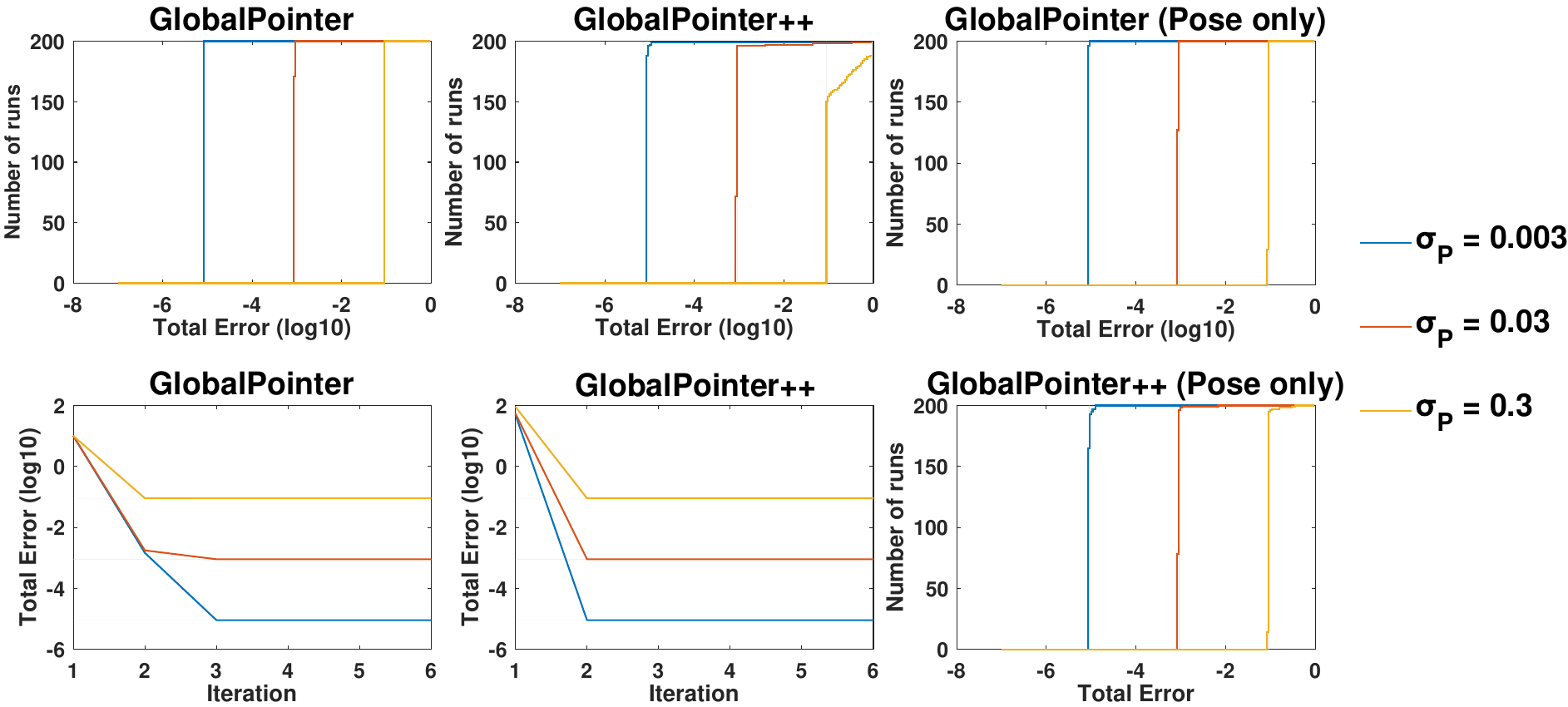}
	\caption{\textbf{Global optimality analysis of our method under three point cloud noise levels.}}
	\label{fig:Global Optimality Check}
\end{figure}

In this section, we validate the global convergence performance of \textit{GlobalPointer}, \textit{GlobalPointer++}, pose-only \textit{GlobalPointer}, and pose-only \textit{GlobalPointer++}. We conduct 200 independent trials under three different point cloud noise levels with random initialization. As shown in \cref{fig:Global Optimality Check}, \textit{GlobalPointer} consistently maintains global convergence across varying noise levels. In contrast, \textit{GlobalPointer++} achieves global convergence only when the point cloud noise is small, as the plane adjustment with plane-to-plane error relies on well-estimated local normal vectors. However, \textit{GlobalPointer++} converges with fewer iterations, demonstrating its efficiency. Similarly, pose-only \textit{GlobalPointer} consistently achieves global convergence, while pose-only \textit{GlobalPointer++} achieves global convergence only under low point cloud noise levels.

\begin{figure}[t]
	\centering
	\begin{subfigure}[b]{\textwidth}
		\centering
		\includegraphics[width=0.85\textwidth]{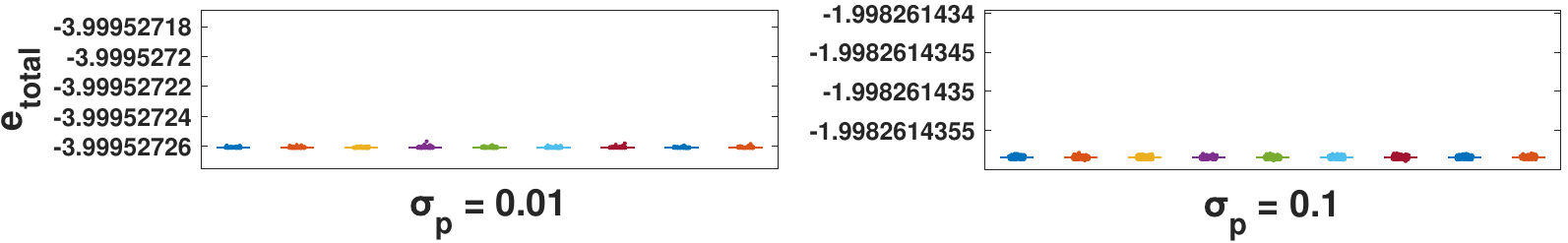}
		\caption{Total error $e_{total}$}
		\label{fig:global profile new:total error}
	\end{subfigure}
	\par\medskip
	\begin{subfigure}[b]{0.46\textwidth}
		\centering
		\includegraphics[width=\textwidth]{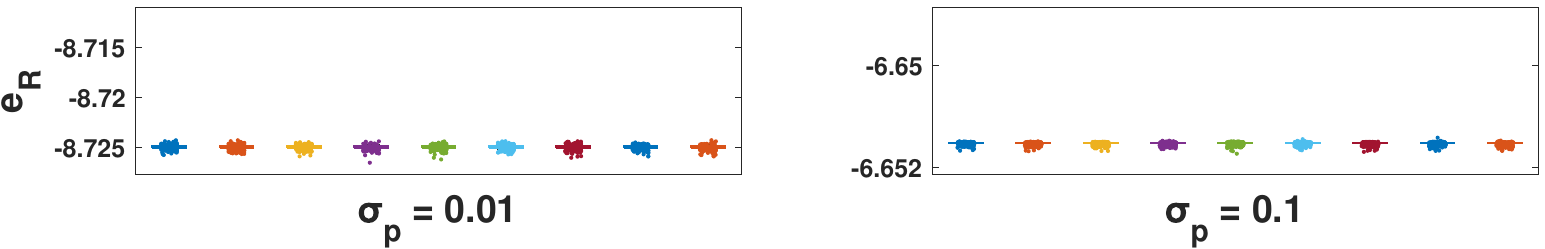}
		\caption{Rotation error $e_R$}
		\label{fig:global profile new:r error}
	\end{subfigure}
	\hfill
	\begin{subfigure}[b]{0.46\textwidth}
		\centering
		\includegraphics[width=\textwidth]{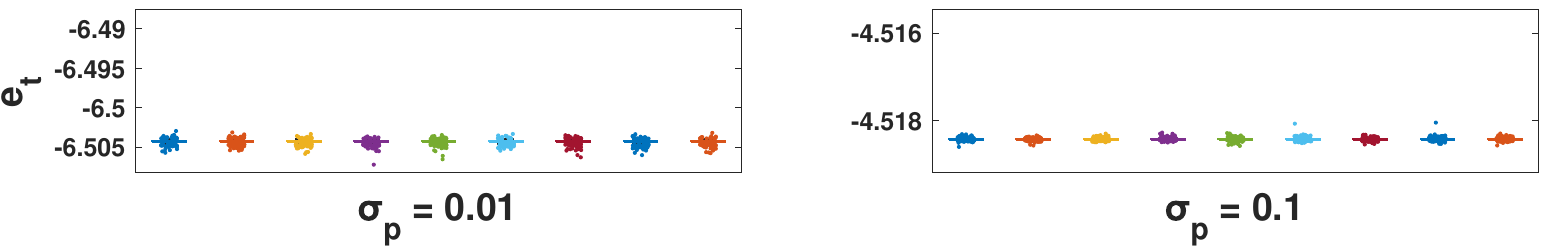}
		\caption{Translation error $e_t$}
		\label{fig:global profile new:t error}
	\end{subfigure}
        \par\medskip
	\begin{subfigure}[b]{0.46\textwidth}
		\centering
		\includegraphics[width=\textwidth]{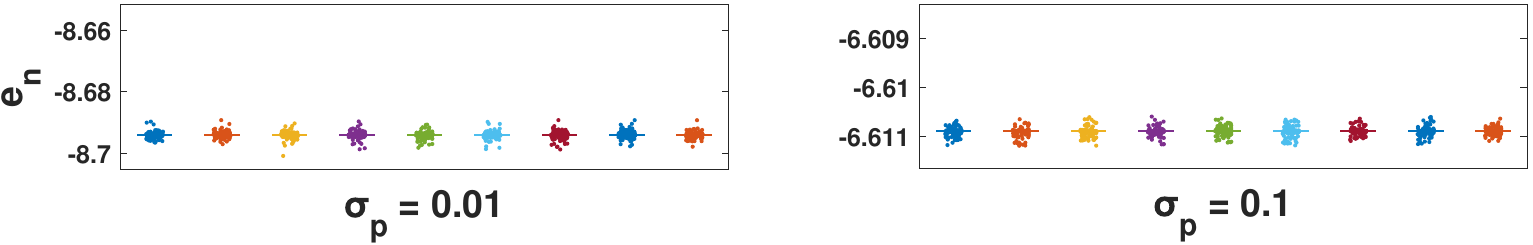}
		\caption{Normal vector error $e_n$}
		\label{fig:global profile new:n error}
	\end{subfigure}
	\hfill
	\begin{subfigure}[b]{0.46\textwidth}
		\centering
		\includegraphics[width=\textwidth]{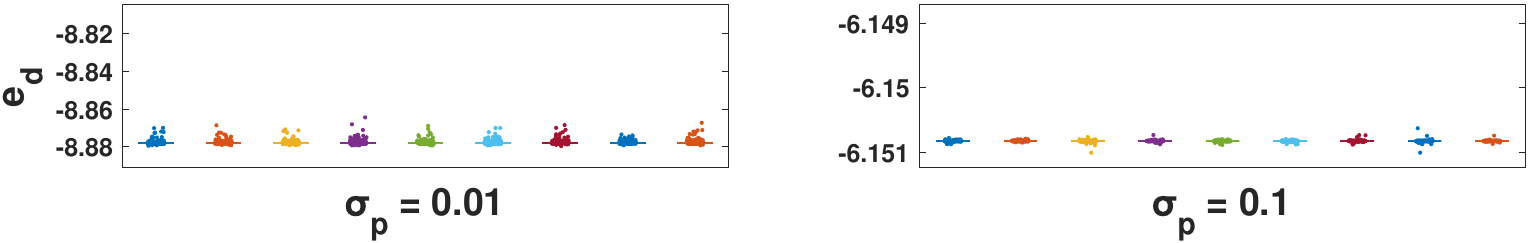}
		\caption{$d$ error $e_d$}
		\label{fig:global profile new:d error}
	\end{subfigure}
	\caption{\textbf{Global convergence analysis of our method on the synthetic dataset.}}
	\label{fig:global profile new}
\end{figure}

To further validate the global convergence of our proposed \textit{GlobalPointer}, we conduct multiple independent trials with random initialization in a synthetic environment consisting of 10 planes and 10 LiDAR poses, fully observed. As shown in \cref{fig:global profile new}, each column box represents a unique random setting. Under two point cloud noise levels, we perform 9 random settings, and for each setting, we run 1000 independent trials with random initialization. The experimental results confirm that our method achieves global optimality even under a high point cloud noise level (${\sigma}_p =0.1$).

\subsection{Synthetic Data Analysis}
\label{supp:subsec:Synthetic Data Analysis}
\begin{figure}[t]
	\centering
	\includegraphics[width=0.99\linewidth]{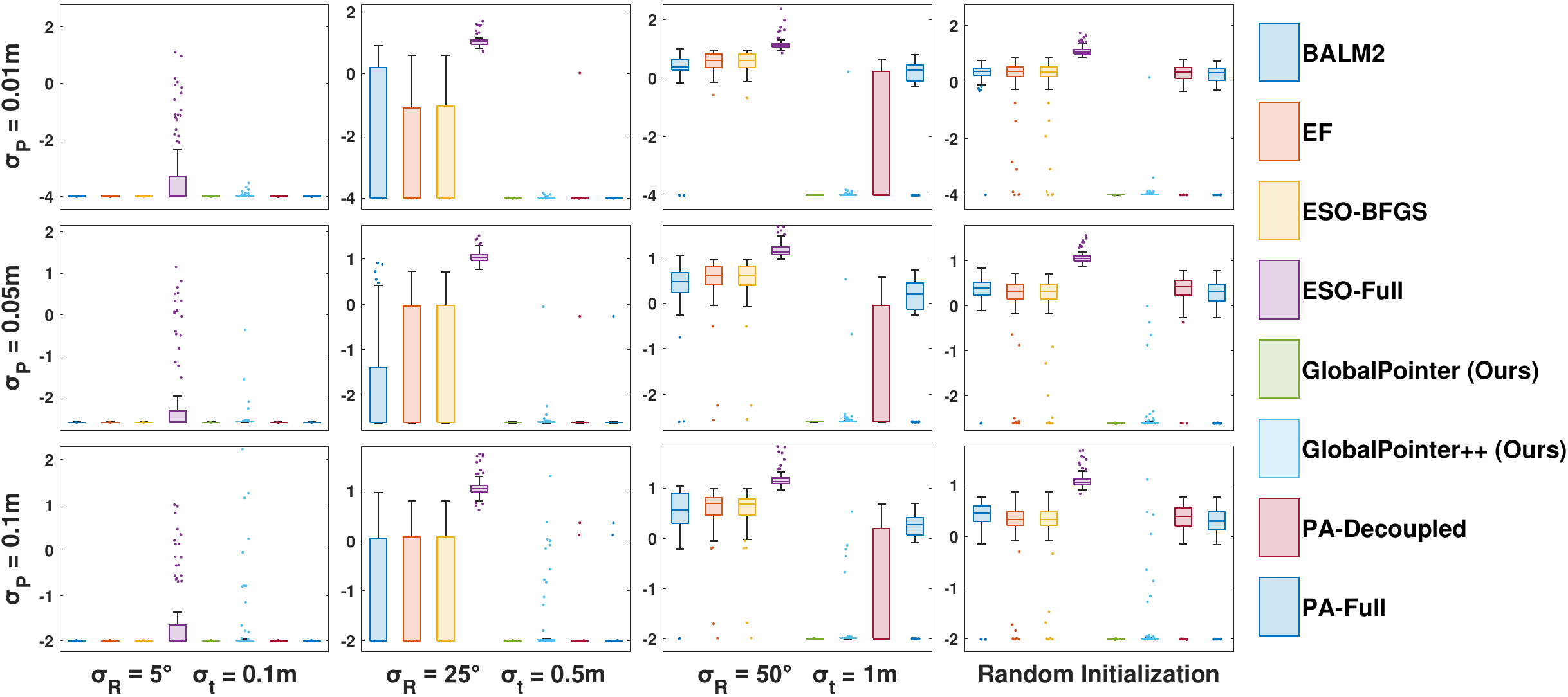}
	\caption{\textbf{Total error $e_{total}$ comparisons on the synthetic data.} The $y$ axis represents the total point-to-plane error $e_{total}$ in the log10 scale.}
\label{fig:synthetic:total error}
\end{figure}

\begin{figure}[t]
	\centering
	\includegraphics[width=0.99\linewidth]{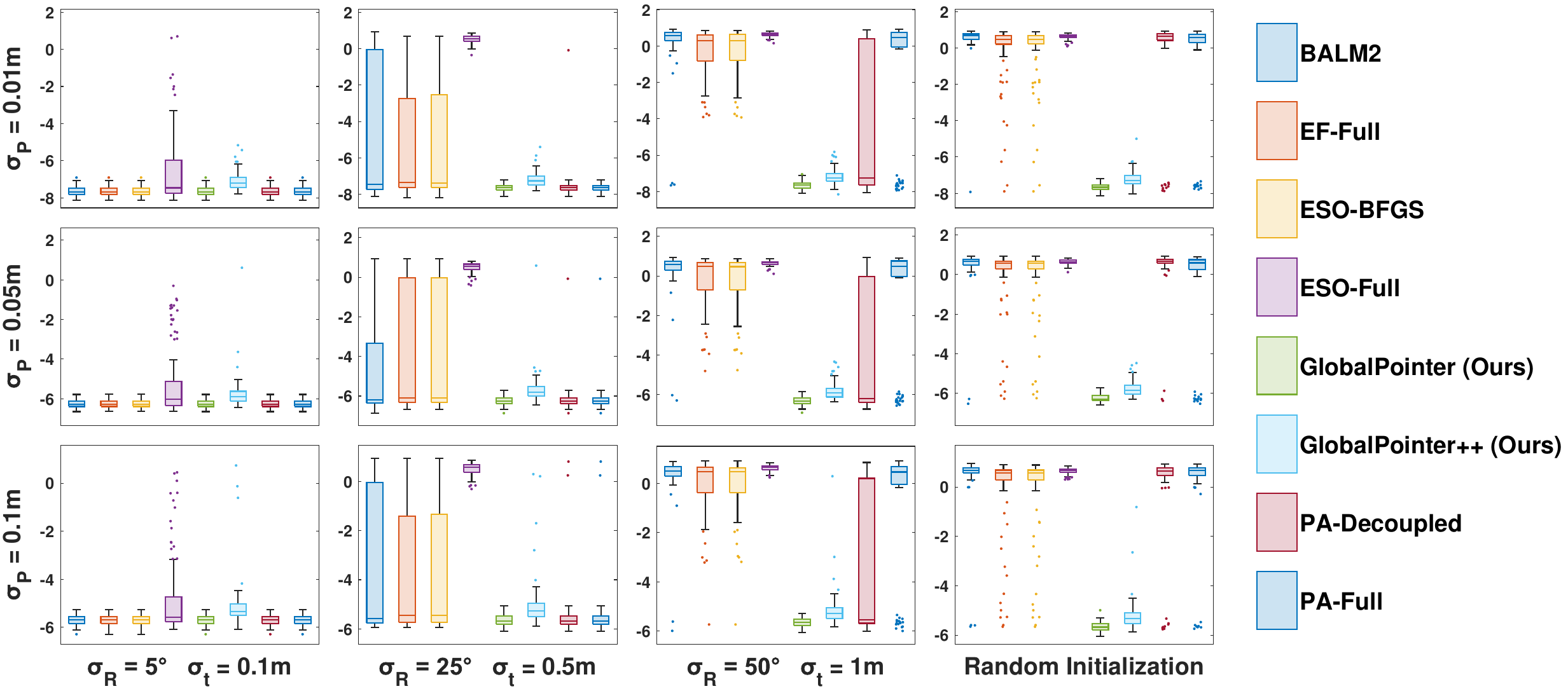}
	\caption{\textbf{Rotation error $e_{R}$ comparisons on the synthetic data.} The $y$ axis represents the total point-to-plane error $e_{total}$ in the log10 scale.}
\label{fig:synthetic:R error}
\end{figure}

\begin{figure}[ht]
	\centering	\includegraphics[width=0.99\linewidth]{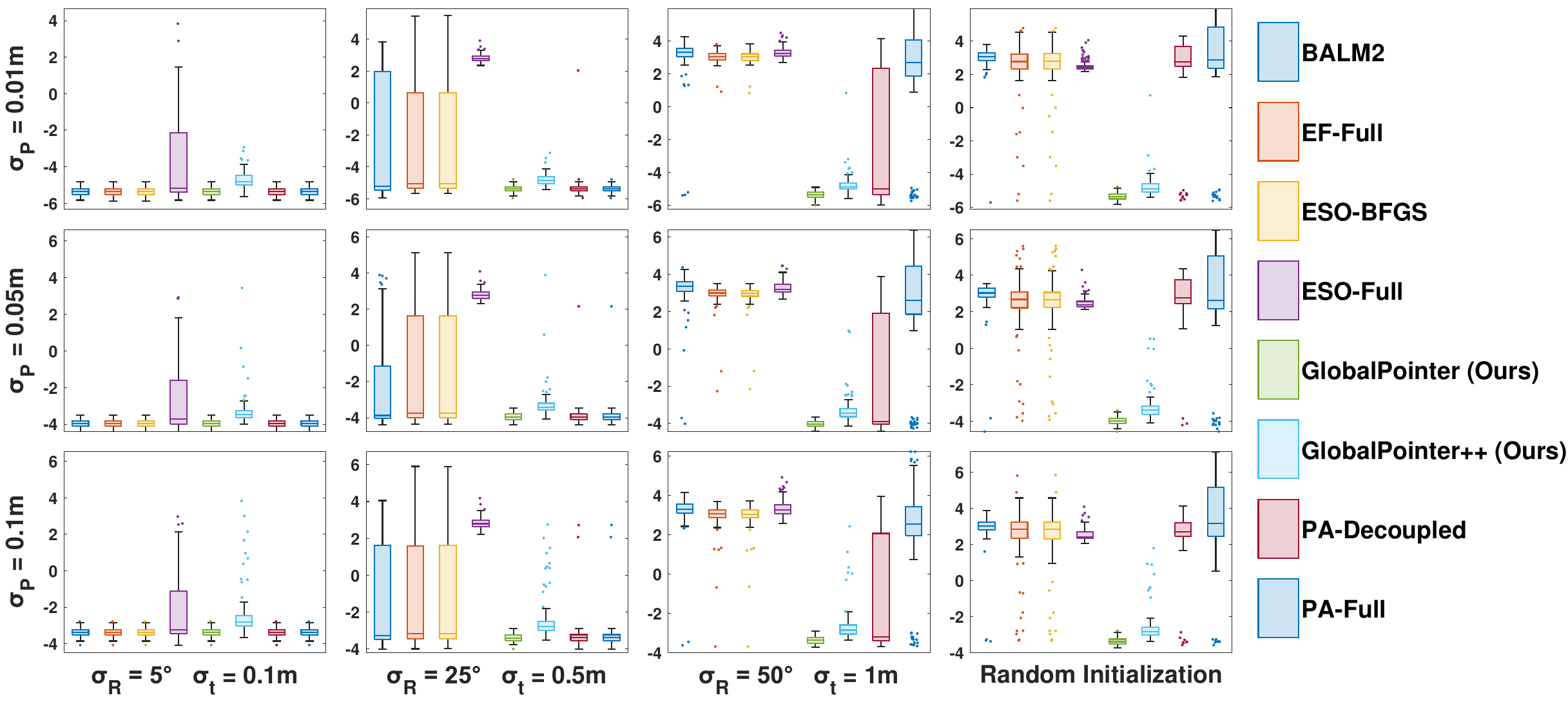}
\caption{\textbf{Translation error $e_{t}$ comparisons on the synthetic data.} The $y$ axis represents the total point-to-plane error $e_{total}$ in the log10 scale.}
\label{fig:synthetic:t error}
\end{figure}

\begin{figure}[ht]
	\centering
	\includegraphics[width=0.99\linewidth]{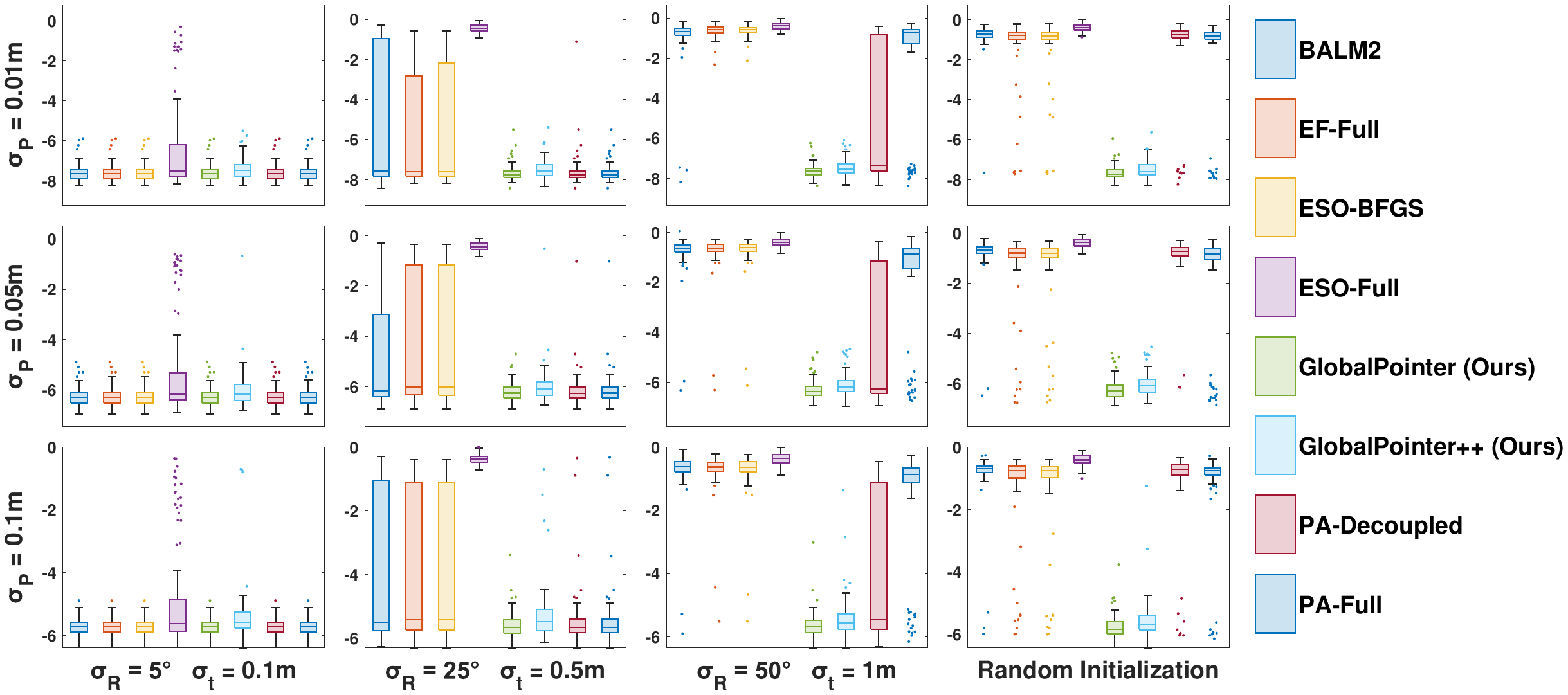}
	\caption{\textbf{Normal vector error $e_{n}$ comparisons on the synthetic data.} The $y$ axis represents the total point-to-plane error $e_{total}$ in the log10 scale.}
\label{fig:synthetic:n error}
\end{figure}

\begin{figure}[ht]
	\centering
	\includegraphics[width=0.99\linewidth]{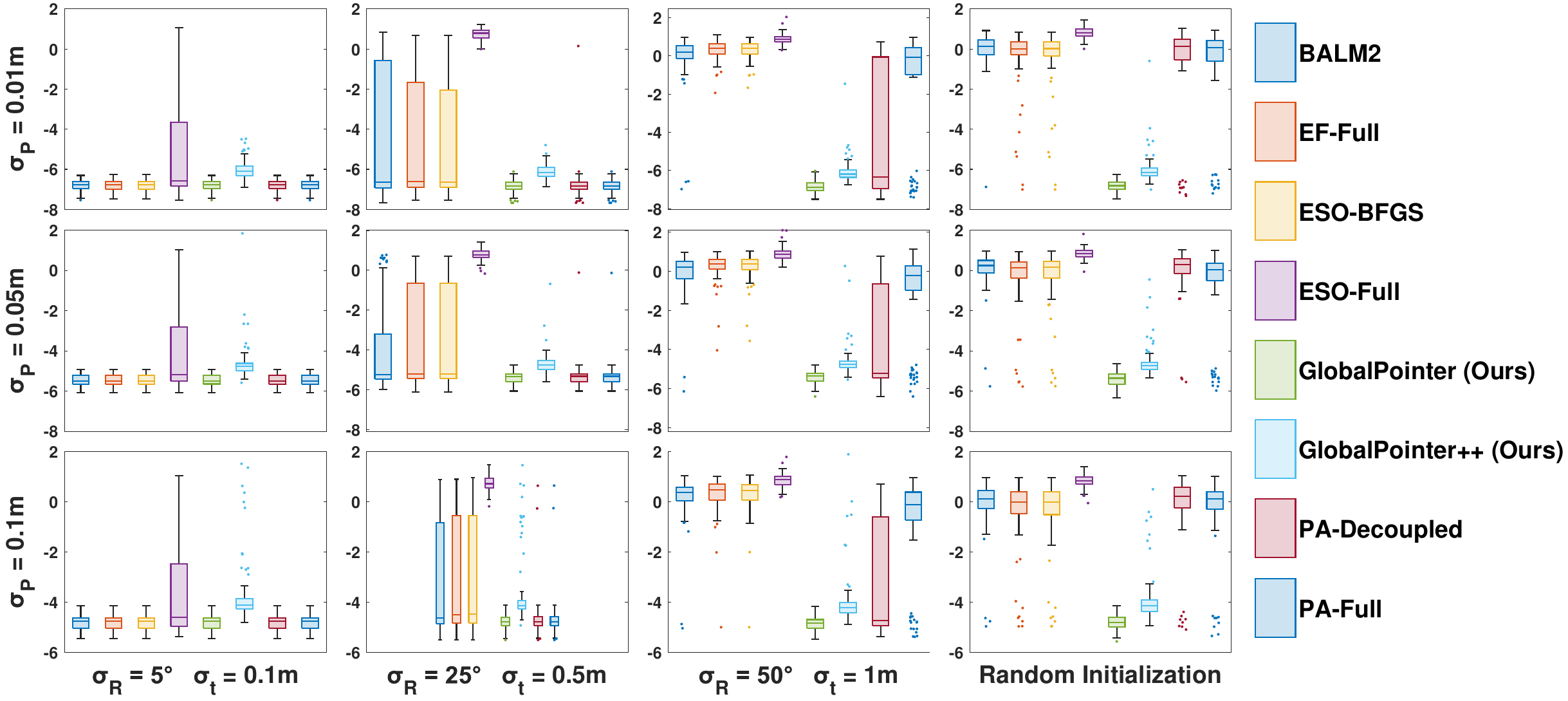}
	\caption{\textbf{$d$ error $e_{d}$ comparisons on the synthetic data.} The $y$ axis represents the total point-to-plane error $e_{total}$ in the log10 scale.}
\label{fig:synthetic:d error}
\end{figure}

We extensively compare our proposed algorithms with other methods in terms of accuracy under varying levels of point cloud noise and pose initialization noise. All statistics are computed over 50 independent trials. As shown in \cref{fig:synthetic:total error,fig:synthetic:R error,fig:synthetic:t error,fig:synthetic:n error,fig:synthetic:d error}, \textit{GlobalPointer} consistently converges to the global optimum across all settings, while \textit{GlobalPointer++} is more sensitive to point cloud noise. ESO \cite{ESO} performs poorly due to unstable numerical stability in Hessian matrix derivation, and other methods fail to converge to the global minimum above a certain level of noise.


\subsection{Real Data Analysis}
\label{supp:subsec:Real Data Analysis}
\begin{figure}
	\centering
	\begin{subfigure}[b]{\textwidth}
		\centering
		\includegraphics[width=0.9\textwidth]{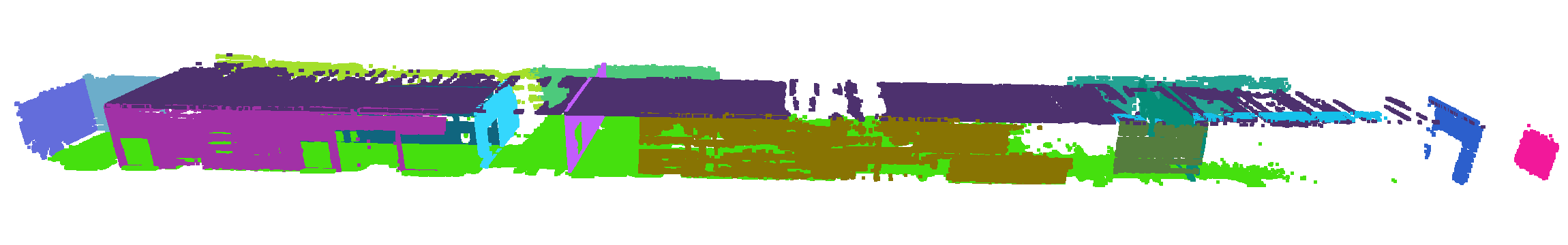}
		\caption{Basement4 Sequence}
		\label{fig:sub1}
	\end{subfigure}
	\par\medskip
	\begin{subfigure}[b]{0.45\textwidth}
		\centering
		\includegraphics[width=\textwidth]{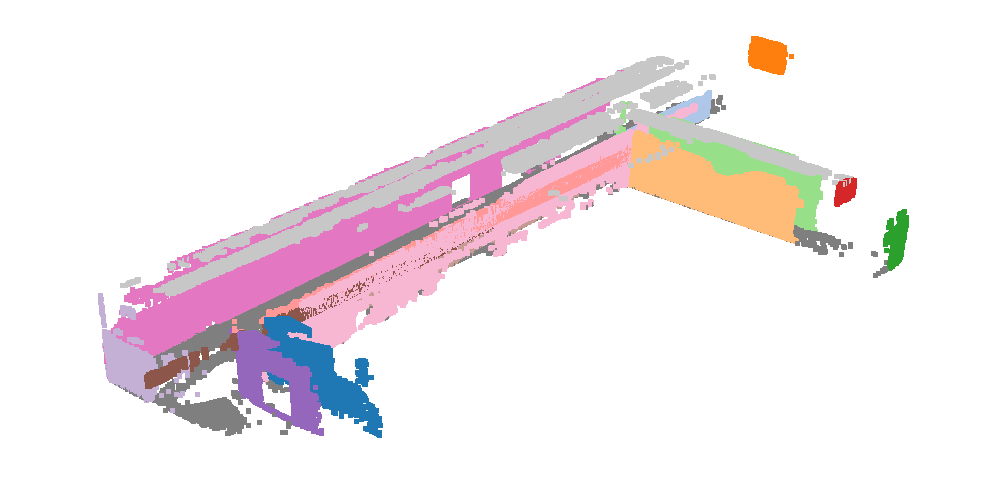}
		\caption{Basement1 Sequence}
		\label{fig:sub2}
	\end{subfigure}
	\hfill
	\begin{subfigure}[b]{0.45\textwidth}
		\centering
		\includegraphics[width=\textwidth]{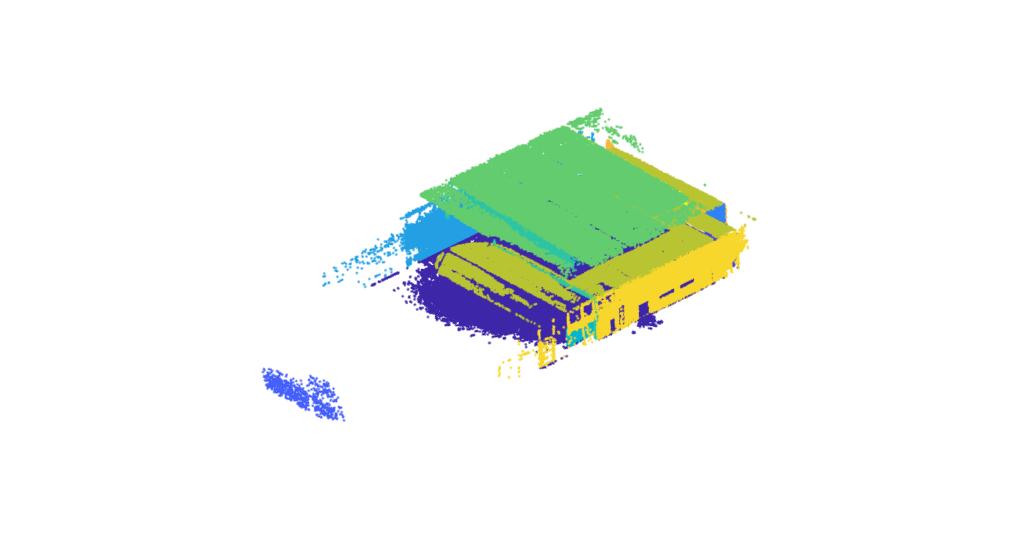}
		\caption{RPG Sequence}
		\label{fig:sub3}
	\end{subfigure}
        \par\medskip
	\begin{subfigure}[b]{0.45\textwidth}
		\centering
		\includegraphics[width=\textwidth]{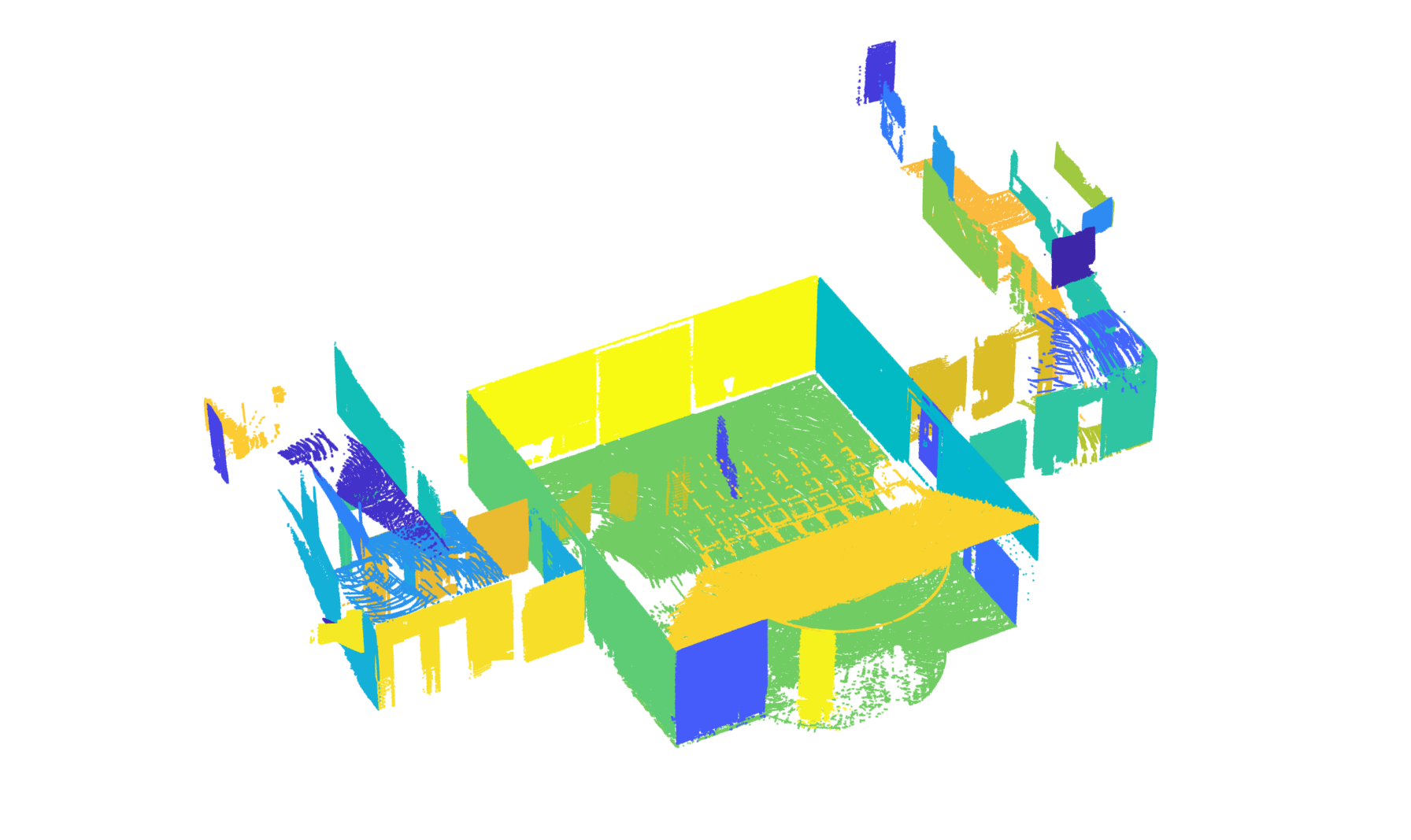}
		\caption{Basement1 Sequence}
		\label{fig:sub2}
	\end{subfigure}
	\hfill
	\begin{subfigure}[b]{0.45\textwidth}
		\centering
		\includegraphics[width=\textwidth]{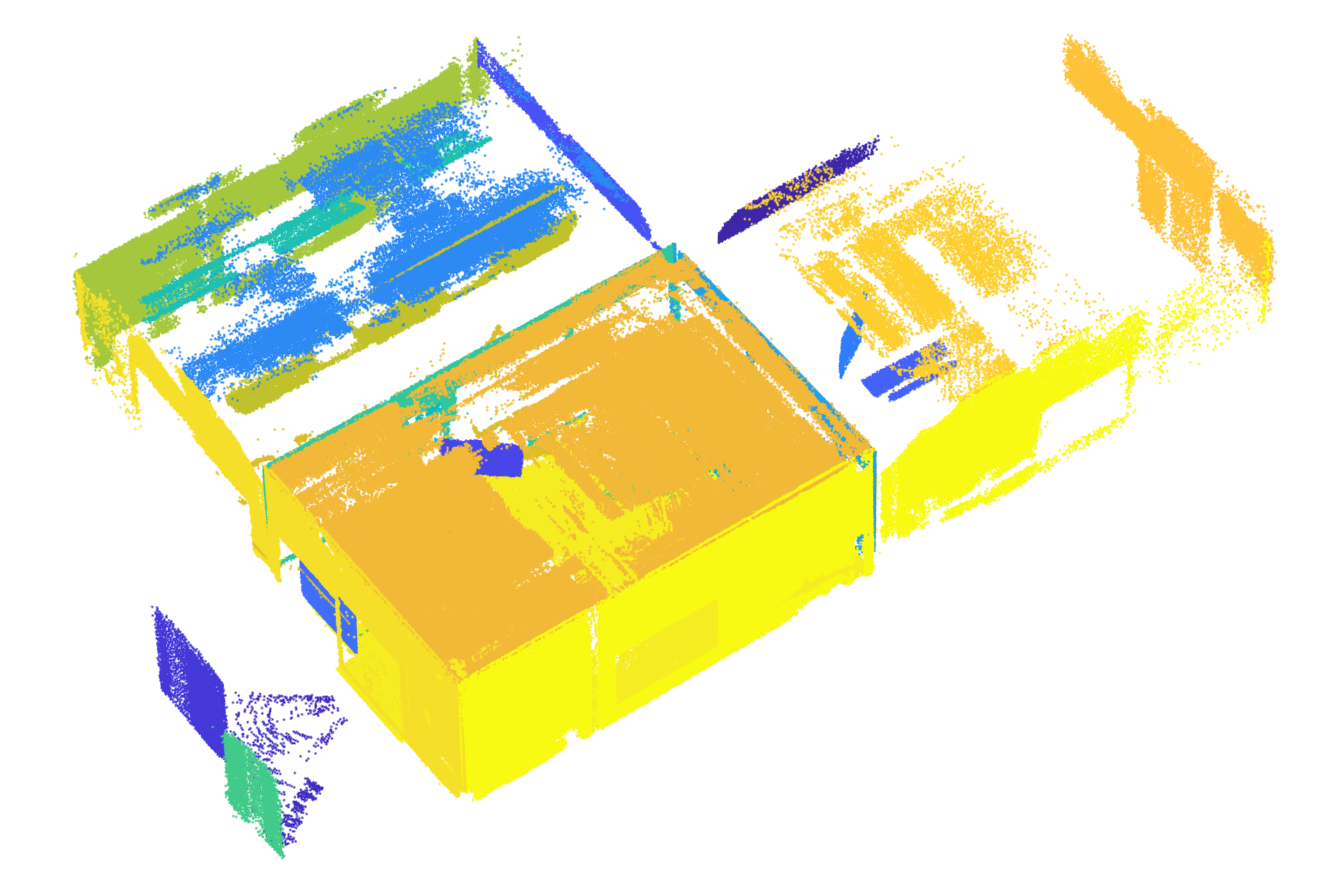}
		\caption{RPG Sequence}
		\label{fig:sub3}
	\end{subfigure}
	\caption{\textbf{Plane segmentation results of the real LiDAR dataset.}}
	\label{fig:segementation results}
\end{figure}

\begin{figure}[ht]
	\centering
	\includegraphics[width=0.99\linewidth]{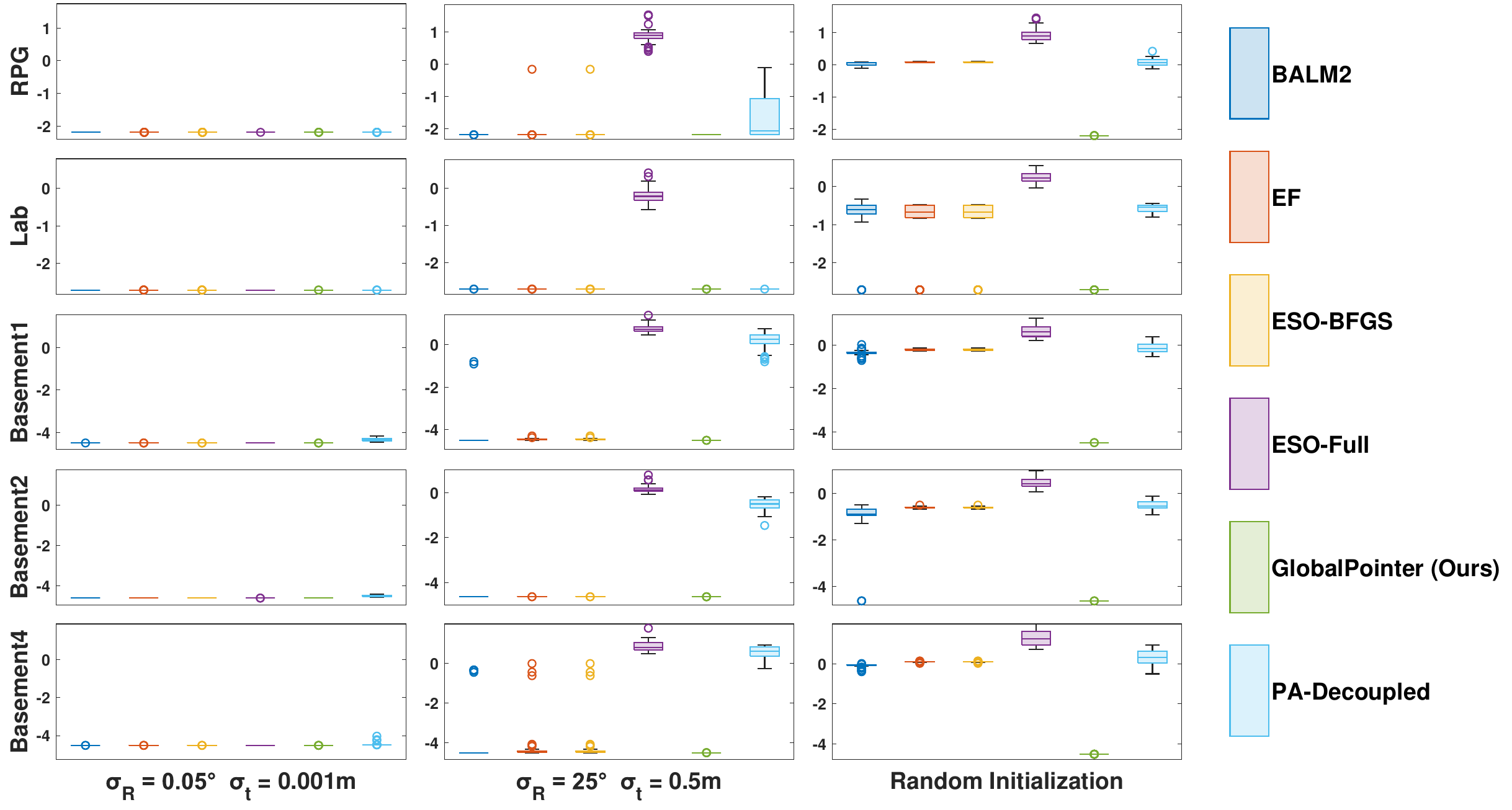}
	\caption{\textbf{Total error $e_{total}$ comparisons on the real dataset.} The $y$ axis represents the total point-to-plane error $e_{total}$ in the log10 scale.}
	\label{fig:real:total error}
\end{figure}

\begin{figure}[ht]
	\centering
	\includegraphics[width=0.99\linewidth]{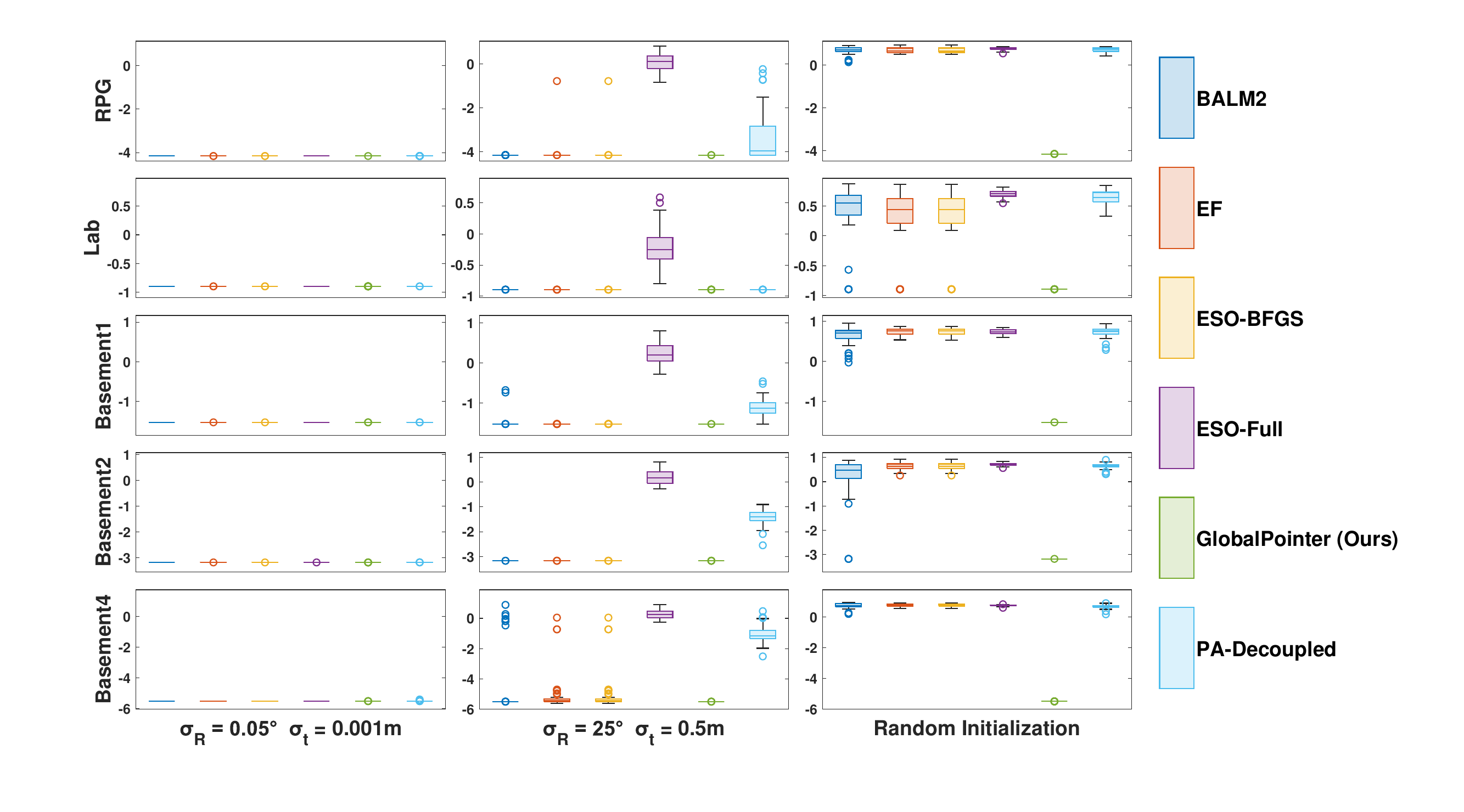}
	\caption{\textbf{Rotation error $e_{R}$ comparisons on the real dataset.} The $y$ axis represents the total point-to-plane error $e_{total}$ in the log10 scale.}
	\label{fig:real:r error}
\end{figure}

\begin{figure}[ht]
	\centering
	\includegraphics[width=0.99\linewidth]{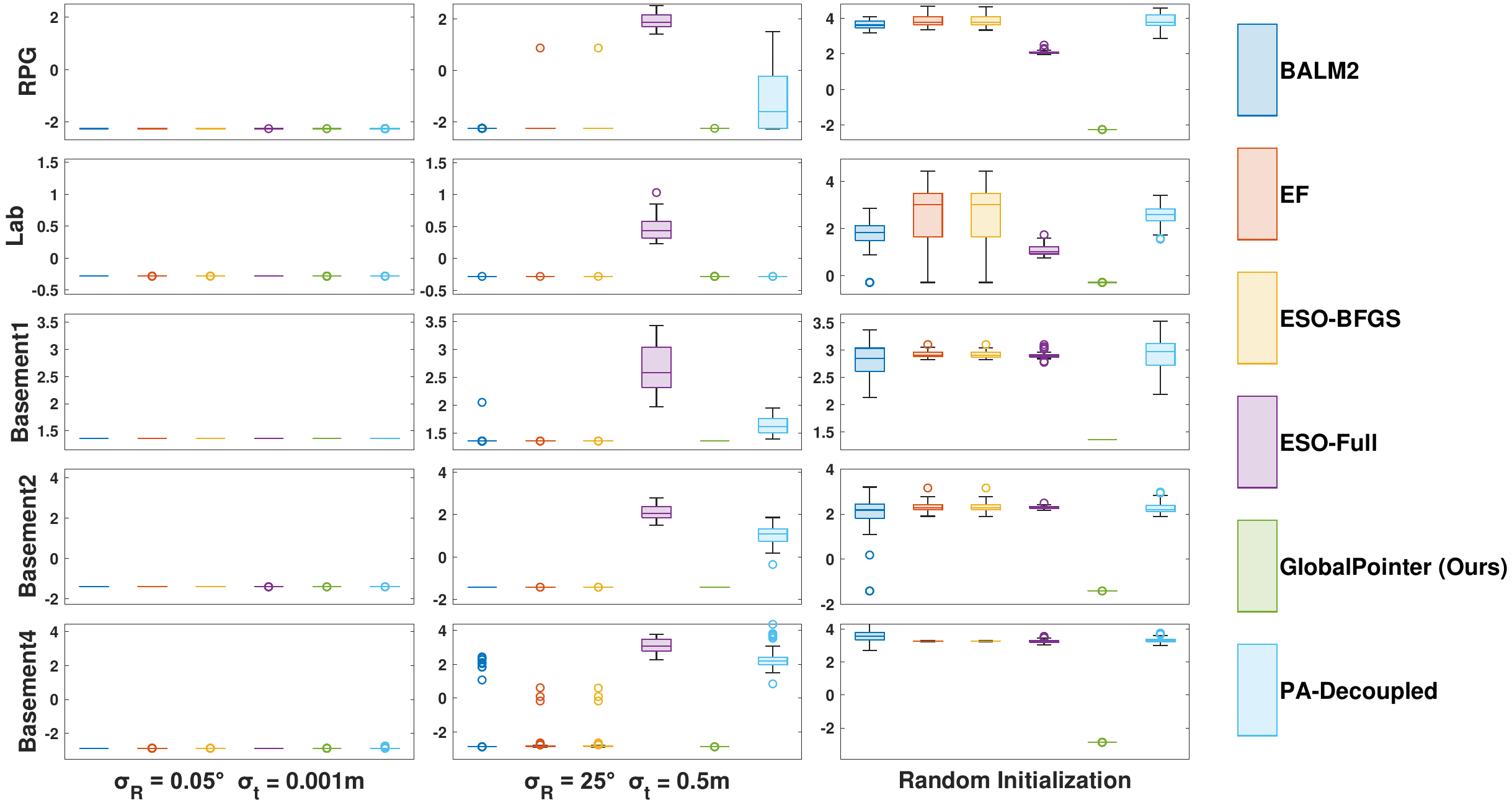}
	\caption{\textbf{Translation error $e_{t}$ comparisons on the real dataset.} The $y$ axis represents the total point-to-plane error $e_{total}$ in the log10 scale.}
	\label{fig:real:t error}
\end{figure}

\begin{figure}[ht]
	\centering
	\includegraphics[width=0.99\linewidth]{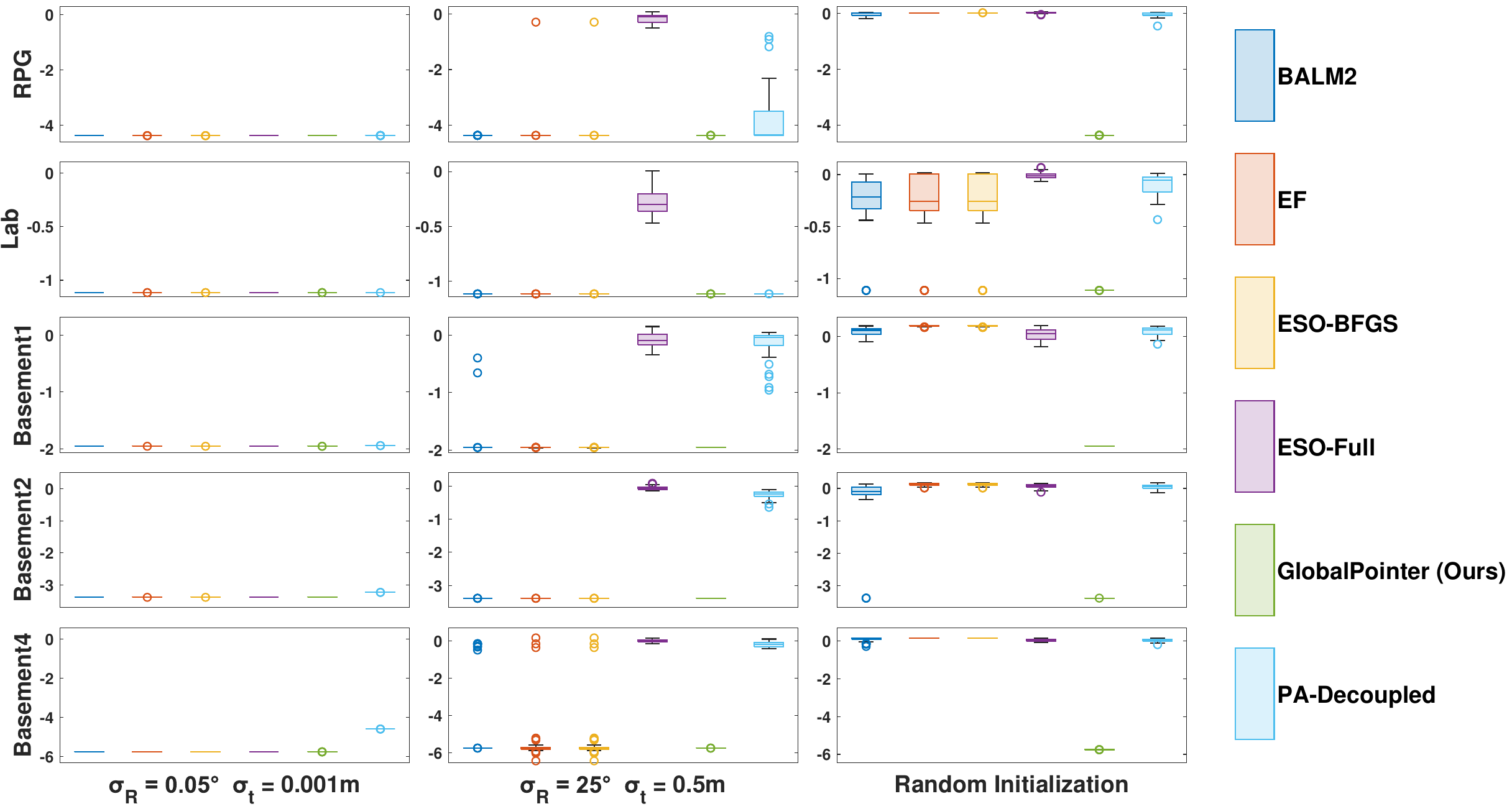}
	\caption{\textbf{Normal vector error $e_{n}$ comparisons on the real dataset.} The $y$ axis represents the total point-to-plane error $e_{total}$ in the log10 scale.}
	\label{fig:real:n error}
\end{figure}

\begin{figure}[ht]
	\centering
	\includegraphics[width=0.99\linewidth]{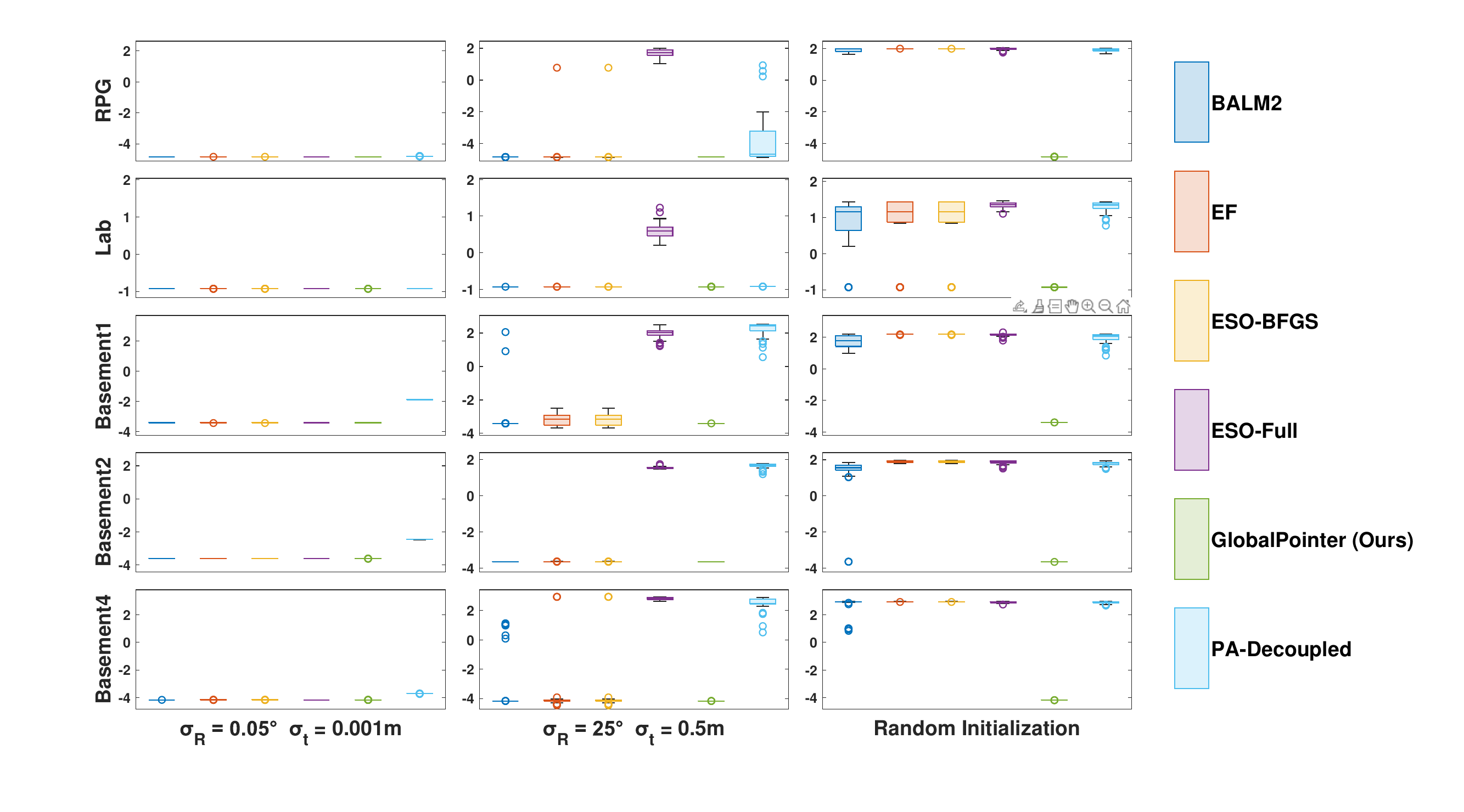}
	\caption{\textbf{$d$ error $e_{d}$ comparisons on the real dataset.} The $y$ axis represents the total point-to-plane error $e_{total}$ in the log10 scale.}
	\label{fig:real:d error}
\end{figure}

For real datasets, we select sequences from the Hilti dataset \cite{helmberger2022hilti} with a higher number of indoor planes. Initially, We transform the local point clouds of each frame into the global coordinate system using ground truth poses. Subsequently, we employ plane segmentation and fitting algorithms to extract point clouds \cite{zhou2018open3d} associated with the same plane, with a RANSAC fitting threshold of $0.01$. The segmentation results are shown in \cref{fig:segementation results}. Furthermore, to align with our hypothesis that each plane should be observed by as many LiDAR frames as possible, we randomly divide all point cloud frames into 50 subsets and accumulate them into individual sub-cloud frames. All statistics are computed over 50 independent trials. In \cref{fig:real:total error,fig:real:r error,fig:real:t error,fig:real:n error,fig:real:d error}, we compare the accuracy of our algorithm with other methods. It is evident that our proposed \textit{GlobalPointer} consistently demonstrates global convergence performance across all settings, while other methods require well-initialized values to converge to the ground truth.


\subsection{Absolute Time Analysis}
\label{supp:subsec:Absolute Time Analysis}
\begin{figure}[ht]
	\centering
	\includegraphics[width=0.99\linewidth]{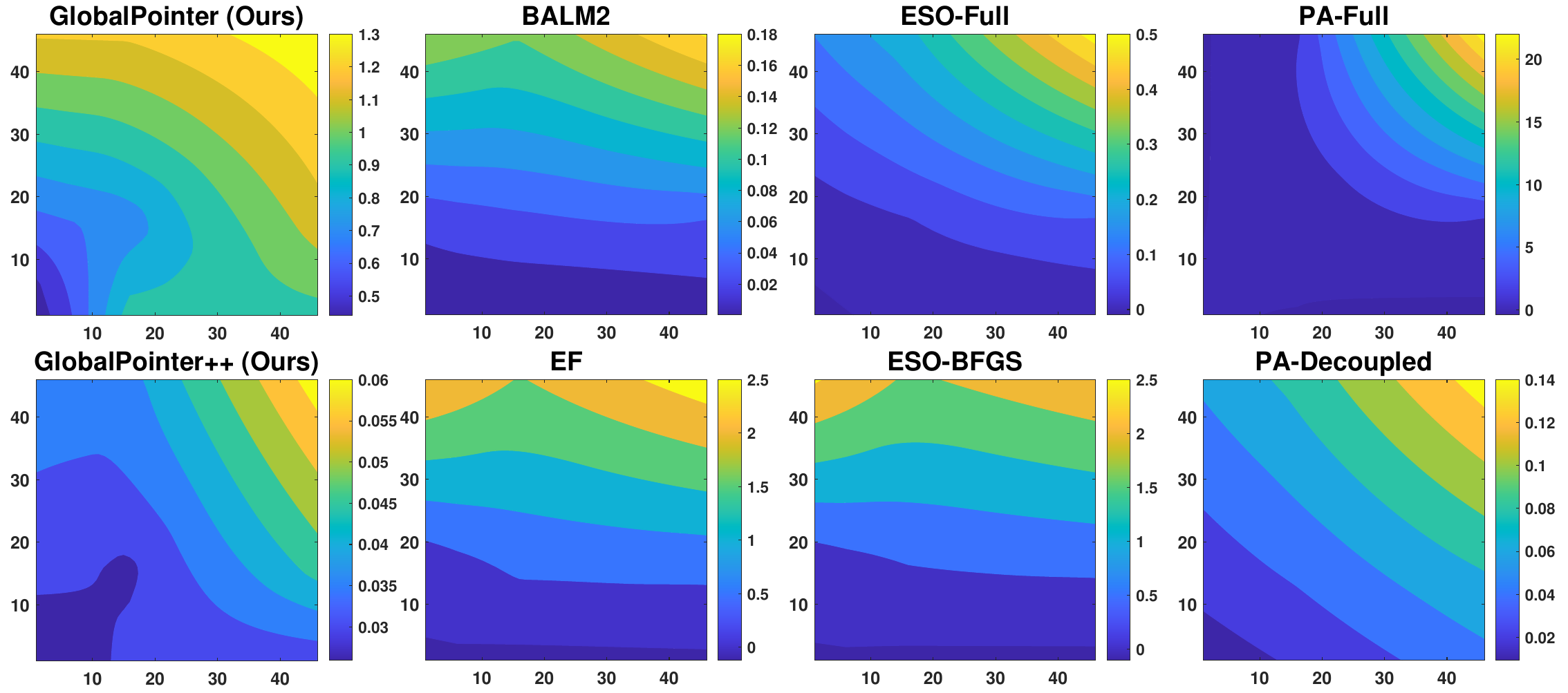}
	\caption{\textbf{Absolute time comparisons.} The $x$ axis represents the number of planes and the $y$ axis represents the number of poses. The color bar on the right side of each subplot indicates the mapping relationship between the multiplier of runtime growth and the colors.}
	\label{fig:absolute time}
\end{figure}

We also evaluate the absolute time of our algorithms. Starting with 5 planes and 5 poses, we incrementally increase the number of poses and planes. We conduct 50 independent trials and use the median time as the metric. Experimental results in \cref{fig:absolute time} demonstrate that the second-order methods BALM2 \cite{BALM2} and ESO-Full \cite{ESO} exhibit superior efficiency, while our proposed \textit{GlobalPointer++} achieves the highest efficiency. In contrast, the first-order methods EF \cite{EF} and ESO-BFGS \cite{ESO} show lower efficiency due to their slower convergence rates. PA-Full \cite{zhou2020efficient} exhibits the worst efficiency because of its cubic time complexity. PA-Decoupled \cite{zhou2020efficient}, which also employs alternating minimization, demonstrates time complexity similar to our method.

\subsection{Overlap Analysis}
\label{supp:subsec:Overlap Analysis}
\begin{figure}[h]
  \centering
  \includegraphics[width=0.99\linewidth]{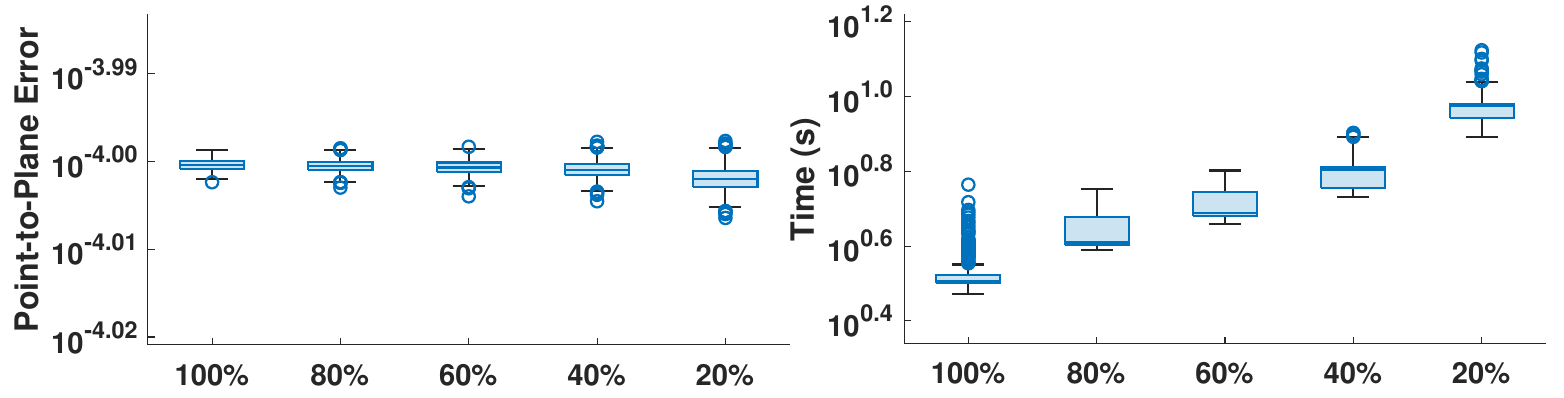}
   \caption{\textbf{Illustration of the point-to-plane error (left) and time (right) with the decreasing overlap ratio given the random poses and planes.}}
   \label{fig:error_time}
\end{figure}

As highlighted in {\it Remark 4}, the overlap between planes and poses is a critical factor for our algorithm's performance. We conduct a study to evaluate the impact of this factor on the accuracy and efficiency of our algorithm. The evaluation is performed by decreasing the overlap ratio from 100\% to 20\% in a synthetic environment comprising 100 planes and 100 LiDAR poses. Random initialization and varying degrees of random overlap are employed for each experiment.  All statistics are computed over 100 independent trials. As shown in \cref{fig:error_time}, our method achieves consistent accuracy across different overlap levels, while requiring more time to converge as overlap decreases.

\clearpage

\bibliographystyle{splncs04}
\bibliography{main}

\end{document}